\documentclass[10pt,journal,compsoc]{IEEEtran}
\usepackage{amsmath,amsfonts}
\usepackage{bm}
\usepackage{algorithm}
\usepackage{algpseudocode}
\usepackage{array}
\usepackage[caption=false,font=normalsize,labelfont=sf,textfont=sf]{subfig}
\usepackage{textcomp}
\usepackage{stfloats} 
\usepackage{url}
\usepackage{verbatim}
\usepackage{graphicx}
\usepackage{cite}
\usepackage{xcolor}
\usepackage[T1]{fontenc}
\usepackage{booktabs}
\usepackage{rotating}
\usepackage{amssymb}
\usepackage{pifont}
\usepackage{wrapfig}
\usepackage{tikz}
\usepackage{hyperref}
\usepackage{multirow}
\usepackage{multicol}
\usepackage{tcolorbox}
\usepackage{enumitem}
\usepackage{subcaption}
\usepackage{ragged2e}
\usepackage{ulem}
\usepackage{color, colortbl}
\definecolor{Gray}{gray}{0.85}

\usepackage{dsfont}

\usepackage{tabularx}

\newcommand*\colourcheck[1]{%
  \expandafter\newcommand\csname #1check\endcsname{\textcolor{#1}{\ding{52}}}%
}
\colourcheck{mygreen}
\colourcheck{myred}
\usepackage{subfig}  % 加载subfig包

\definecolor{mygreen}{rgb}{0.0, 0.47, 0.0}
\definecolor{myred}{rgb}{0.85, 0.0, 0.0}
\definecolor{myblue}{rgb}{0.0, 0.0, 0.85}
\definecolor{deepcarrotorange}{rgb}{0.91, 0.41, 0.17}
\definecolor{amber}{rgb}{0.95, 0.60, 0.0}
\definecolor{amaranth}{rgb}{0.9, 0.17, 0.31}

\def\halfcheckmark{\tikz\draw[scale=0.2,fill=amber](0,.35) -- (.25,0) -- (1,.7) -- (.25,.15) -- cycle (0.75,0.2) -- (0.77,0.2)  -- (0.6,0.7) -- cycle;}
\newcommand{\xmark}{\ding{55}}%
\newcommand{\increase}[1]{\textcolor{mygreen}{\small{(+#1)}}}

\newcommand{\pxj}[1]{\textcolor{black}{#1}}

\makeatletter
\newcommand{\thickhline}{%
    \noalign {\ifnum 0=`}\fi \hrule height 1pt
    \futurelet \reserved@a \@xhline
}
\makeatother

\begin{document}

\title{Emotion-LLaMAv2 and MMEVerse: \\A New Framework and Benchmark for \\Multimodal Emotion Understanding}

% \author{
% Xiaojiang~Peng\textsuperscript{1},
% Jingyi~Chen\textsuperscript{1},
% Zebang~Cheng\textsuperscript{1},\\
% Fengyi~Wu\textsuperscript{2},
% Yifei~Dong\textsuperscript{2},
% Shuyuan~Tu\textsuperscript{2},
% Qiyu~Hu\textsuperscript{2},
% Huiting~Huang\textsuperscript{2},\\
% Yuxiang~Lin\textsuperscript{1},
% Jun{-}Yan~He\textsuperscript{2},
% Kai~Wang\textsuperscript{3},
% Zheng~Lian \textsuperscript{4},
% Jingdong~Sun\textsuperscript{5},\\
% Alexander~G.~Hauptmann\textsuperscript{5},
% Zhi{-}Qi~Cheng\textsuperscript{2}\thanks{Corresponding author.}
% }

% \date{
% \textsuperscript{1}Shenzhen Technology University \\
% \textsuperscript{2}University of Washington \\
% \textsuperscript{3}National University of Singapore \\
% \textsuperscript{4}Institute of Automation, Chinese Academy of Sciences \\
% \textsuperscript{5}Carnegie Mellon University
% }

\markboth{Journal of \LaTeX\ Class Files,~Vol.~14, No.~8, August~2021}%
{Shell \MakeLowercase{\textit{et al.}}: A Sample Article Using IEEEtran.cls for IEEE Journals}
\author{
Xiaojiang~Peng, \textit{Senior Member, IEEE},
Jingyi~Chen,
Zebang~Cheng,
Bao~Peng,
Fengyi~Wu,
Yifei~Dong,
Shuyuan~Tu,
Qiyu~Hu,
Huiting~Huang,
Yuxiang~Lin,
Jun{-}Yan~He,
Kai~Wang,
Zheng~Lian, \textit{Senior Member, IEEE},
Zhi{-}Qi~Cheng, \textit{Senior Member, IEEE}%
\thanks{Corresponding author (zhiqics@uw.edu).}
\IEEEcompsocitemizethanks{
\IEEEcompsocthanksitem
This work was primarily conducted by the teams at Shenzhen Technology University and the University of Washington.

\IEEEcompsocthanksitem
Xiaojiang~Peng, Jingyi~Chen, Zebang~Cheng, and Yuxiang~Lin are with Shenzhen Technology University. Bao~Peng is with Shenzhen University of Information Technology.

\IEEEcompsocthanksitem
Fengyi~Wu, Yifei~Dong, Qiyu~Hu, Shuyuan~Tu, Huiting~Huang, and Zhi{-}Qi~Cheng are with the University of Washington.

\IEEEcompsocthanksitem
Jun{-}Yan~He is with the Foundation Model Team, Meituan.

\IEEEcompsocthanksitem
Kai~Wang is with the National University of Singapore.

\IEEEcompsocthanksitem
Zheng~Lian is with Tongji University.
}
}

\IEEEtitleabstractindextext{
\begin{abstract}
\justifying
Understanding human emotions from multimodal signals poses a significant challenge in affective computing and human-robot interaction. While multimodal large language models (MLLMs) have excelled in general vision-language tasks, their capabilities in emotional reasoning remain limited. The field currently suffers from a scarcity of large-scale datasets with high-quality, descriptive emotion annotations and lacks standardized benchmarks for evaluation.
% Existing methods often rely on handcrafted facial priors, lack large-scale instruction tuning data, and provide no standardized benchmark for evaluation. 
Our preliminary framework, Emotion-LLaMA, pioneered instruction-tuned multimodal learning for emotion reasoning but was restricted by explicit face detectors, \pxj{implicit} fusion strategies, and \pxj{low-quality training data with limited scale}. To address these limitations, we present Emotion-LLaMAv2 and the MMEVerse benchmark, establishing an \pxj{end-to-end pipeline together with a standardized evaluation setting for emotion recognition and reasoning.}
%raw输入有点和end to end重复\textcolor{red}{end to end pipeline from raw multimodal inputs to emotion reasoning outputs together with a standardized evaluation setting} for multimodal emotion understanding. 
Emotion-LLaMAv2 introduces three key advances. 
First, an end-to-end multiview encoder eliminates external face detection and \pxj{captures nuanced emotional cues via richer spatial and temporal multiview tokens.}
% \textcolor{red}{implicit spatial and temporal focusing} through cross modal attention. 第一个点主要是把v1的pooled表达改为seq 的
Second, \pxj{a Conv Attention pre-fusion module is designed to enable simultaneous local and global multimodal feature interactions external to the LLM backbone.}
%   nuanced temporal emotional cues via  \textcolor{red}{without modifying the LLM backbone}. 几乎所有做工作都不改变LLM
Third, \pxj{a perception-to-cognition} curriculum instruction tuning scheme within the LLaMA2 backbone unifies emotion recognition and free-form emotion reasoning. 
To support large-scale training and reproducible evaluation, MMEVerse aggregates twelve publicly available emotion datasets, including IEMOCAP, MELD, DFEW, and MAFW, into a unified multimodal instruction format. The data are re-annotated via a multi-agent pipeline involving Qwen2 Audio, Qwen2.5 VL, and GPT 4o, producing 130k training clips and 36k testing clips across 18 evaluation benchmarks. Experiments demonstrate that Emotion-LLaMAv2 consistently outperforms representative open source MLLMs such as Qwen2.5 Omni and AffectGPT, achieving state-of-the-art performance on MER-UniBench and MMEVerse-Bench, with improved generalization and more structured multimodal reasoning behavior. 
Overall, Emotion-LLaMAv2 and MMEVerse provide a reproducible research foundation for studying multimodal emotion recognition and reasoning, and represent a step toward connecting low-level multimodal perception with emotion-centered language reasoning. Our code and benchmark are available at \href{https://github.com/ooochen-30/Emotion-LLaMA-v2}{\textsf{https://github.com/ooochen-30/Emotion-LLaMA-v2}}.
\end{abstract}

}
% }

\maketitle

\section{Introduction}
\label{sec:intro}

\IEEEPARstart{U}{nderstanding} human emotions is a central challenge in affective computing and human-robot interaction~\cite{cheng2016video, mackenzie2024human, imani2019survey, cao2020psychological}. Accurate emotion understanding enables artificial agents to perceive and adapt to users’ affective states, fostering empathy and effective communication in domains such as education, healthcare, robotics, and counseling. 
In computational settings, emotion understanding extends beyond categorical prediction and requires models to jointly recognize emotional states and reason about their underlying causes and contextual expressions. Unlike conventional recognition tasks, emotion understanding integrates perception, cognition, and context, making it inherently complex. Human affect emerges from multimodal signals, including speech prosody, facial micro expressions, and situational context, which are asynchronous, weakly correlated, and shaped by latent psychological and environmental factors. 
Despite decades of progress in computer vision, speech processing, and natural language understanding, achieving robust and generalizable emotion understanding that faithfully captures the subtle, dynamic, and context-dependent nature of human affect from raw multimodal observations remains an open problem.

Early studies on emotion analysis primarily focused on unimodal recognition, including facial expression analysis~\cite{jiang2020dfew, li2024two, ngwe2023patt, wang2020suppressing}, speech-based emotion detection~\cite{fan2021lssed, hsu2021hubert}, and textual sentiment analysis~\cite{devlin2018bert, lei2023instructerc, hung2023beyond}. While effective within individual modalities, these approaches fail to robustly capture the holistic and interdependent nature of human affect. 
To mitigate this limitation, subsequent multimodal fusion frameworks~\cite{cheng2023semi, li2024mm, zadeh2018multimodal, zhang2023learning} introduced cross-modal attention and joint representation learning, leading to improved recognition performance under controlled experimental settings. However, most existing multimodal frameworks remain limited to modeling feature-level correlations across modalities and do not explicitly support unified semantic representations or emotion-centric reasoning processes. 
In practice, human affective cognition involves hierarchical inference that connects low-level perceptual cues, such as facial movements or vocal intonation, to higher-level appraisal, interpretation, and explanation of emotional states. This gap between computational models and human affective reasoning highlights the need for a unified framework that integrates multimodal perception, semantic abstraction, and explicit emotion reasoning.

Recent advances in multimodal large language models (MLLMs)~\cite{lei2024large, liu2024visual, zhu2023minigpt, guo2024stimuvar, lin2023video} enable unified modeling by aligning visual, textual, and auditory modalities within language-centric architectures. Through large-scale pretraining and instruction tuning, these models demonstrate strong generalization across a wide range of open-domain vision language tasks. 
However, their effectiveness in emotion understanding remains limited due to fundamental mismatches between existing MLLM design assumptions and the characteristics of affective signals. In particular, current MLLMs exhibit restricted capability to directly ingest raw audio streams that encode prosodic and paralinguistic information, as well as insufficient sensitivity to subtle and fine-grained visual clues such as micro expressions and blended affective states. 
These limitations are further exacerbated by the scarcity of large-scale emotion-specific instruction data that jointly couples multimodal perception with emotion-centric reasoning objectives, together with the lack of standardized benchmarks for consistent and reproducible evaluation of multimodal emotion reasoning across datasets. As a result, existing MLLMs often struggle to move beyond surface-level correlations toward robust emotion understanding in realistic settings.

Our previous work, Emotion-LLaMA~\cite{cheng2024emotion}, pioneered instruction-tuned multimodal large language models for emotion understanding by introducing the Multimodal Emotion Recognition and Reasoning dataset and the first emotion-specific instruction tuning framework. This study represented an important step toward unifying emotion recognition and emotion reasoning within a single language-centric architecture. 
However, several limitations constrained its scalability and general applicability. Specifically, Emotion-LLaMA relied on explicit facial region extraction using OpenFace~\cite{baltruvsaitis2016openface}, which prevented fully end-to-end optimization and introduced error propagation across processing stages. In addition, the model compressed audio \pxj{or} visual representations into \pxj{pooled individual tokens,}
% static tokens, 是pool后的token 没法说是static
thereby discarding fine-grained temporal dynamics and acoustic variations that are critical for emotion perception. Furthermore, Emotion-LLaMA was trained on the MERR dataset derived from MER2023~\cite{lian2023mer}, which remained limited in scale, coverage, and annotation consistency across emotion categories. 
As a result, Emotion-LLaMA was unable to fully capture temporal acoustic interactions or complex multimodal dependencies required for robust emotion reasoning, highlighting the need for a more scalable end-to-end framework.

To address the aforementioned limitations, we present \textbf{Emotion-LLaMAv2} together with the accompanying \textbf{MultiModal Emotion uniVerse (MMEVerse)} benchmark, forming a unified end-to-end framework and a large-scale resource for multimodal emotion understanding. Emotion-LLaMAv2 is guided by three core design principles. 
First, we develop an end-to-end multi-view encoding architecture that eliminates the reliance on explicit face detectors, allowing emotion-relevant spatial and temporal regions to be implicitly emphasized through cross-modal attention within the LLM backbone. Moreover, the multi-view encoder captures nuanced emotional cues through adaptively pooled spatial and temporal multi-view embeddings. Second, we introduce a Conv Attention pre-fusion module \pxj{that incorporates spatiotemporal visual tokens and sequential acoustic tokens, enabling simultaneous local and global multimodal feature interactions before inputting to the LLM backbone}.
% \textcolor{red}{dynamic temporal and acoustic token modeling},
% enabling the model to preserve fine grained expressive variations across audio and visual streams. 
Third, we adopt a perception-to-cognition curriculum instruction tuning paradigm within the LLaMA2~\cite{touvron2023llama} backbone to jointly optimize categorical emotion recognition and contextual emotion reasoning. 
Together, these design choices enable a fully differentiable multimodal learning pipeline that supports both perception-driven representation learning and language-based emotion reasoning. By explicitly modeling temporal and acoustic dynamics and aligning multimodal representations with language-centric reasoning objectives, Emotion LLaMAv2 achieves improved generalization and robustness in multimodal emotion perception and reasoning.

To support large-scale instruction tuning and reproducible evaluation, we construct \textbf{MMEVerse}, a unified multimodal resource that consolidates twelve existing datasets, including IEMOCAP~\cite{busso2008iemocap}, MELD~\cite{poria2018meld}, DFEW~\cite{jiang2020dfew}, and MAFW~\cite{liu2022mafw}, into a consistent format with temporally aligned visual, auditory, and textual streams. 
To improve annotation quality and semantic coherence across heterogeneous data sources, we employ a multi-agent annotation pipeline that leverages Qwen2 Audio~\cite{chu2024qwen2}, Qwen2.5 VL~\cite{bai2025qwen2}, and GPT 4o~\cite{hurst2024gpt} to generate fine-grained multimodal descriptions grounded in observable audio and visual evidence. 
The resulting corpus comprises approximately 130K clips for training and 36K clips for testing, each accompanied by structured multimodal annotations designed for instruction-based learning and evaluation. Compared with the earlier MERR dataset, MMEVerse more than triples the data scale and further introduces \textbf{MMEVerse Bench}, a comprehensive evaluation suite spanning 18 benchmarks across emotion recognition, sentiment analysis, affective intent prediction, and multimodal reasoning. 
Together, these components establish MMEVerse as a scalable and standardized benchmark that supports consistent comparison and reproducible evaluation for advancing research in multimodal emotion understanding.

% Extensive experiments demonstrate that \textbf{Emotion-LLaMAv2} consistently surpasses prior approaches across both emotion recognition and reasoning benchmarks. Moreover, \textbf{Emotion-LLaMAv2} achieves state-of-the-art performance on MER-UniBench and MMEVerse-Bench, outperforming general-purpose multimodal large language models such as Qwen2.5-Omni, HumanOmni, and AffectGPT. Beyond quantitative gains, the model further exhibits enhanced interpretability, improved cross-domain generalization, and robustness under modality degradation, thereby establishing a strong foundation for scalable and trustworthy affective intelligence.

% Extensive experiments show that \textbf{Emotion-LLaMAv2} consistently improves performance over prior approaches across both emotion recognition and emotion reasoning benchmarks. On MER-UniBench and MMEVerse-Bench, the proposed model achieves competitive results compared with recent general-purpose multimodal large language models, including Qwen2.5-Omni, HumanOmni, and AffectGPT. Beyond overall performance, Emotion-LLaMAv2 demonstrates improved interpretability, stronger cross-domain generalization, and increased robustness under modality degradation, indicating its effectiveness for practical multimodal emotion understanding.

Extensive experiments demonstrate that \textbf{Emotion LLaMAv2} consistently improves performance over prior approaches across both emotion recognition and emotion reasoning benchmarks. On MER UniBench and MMEVerse Bench, the proposed model achieves competitive or superior results compared with recent general purpose multimodal large language models, including Qwen2.5 Omni, HumanOmni, and AffectGPT. 
Beyond overall accuracy, Emotion LLaMAv2 exhibits more structured multimodal reasoning behavior,  
\pxj{enhanced zero-shot capability for emotion reasoning},
% cross domain generalization,
and increased robustness \pxj{in recognizing basic emotions, sentiment, intention, etc.}
%under modality degradation, \textcolor{red}{as evidenced by consistent performance gains across diverse datasets and missing modality settings}. 
These results indicate the effectiveness of Emotion-LLaMAv2 for practical multimodal emotion perception and reasoning.

\textbf{In summary, our main contributions can be concluded as follows:}
\begin{itemize}[leftmargin=6mm]
    \item \textbf{Unified Dataset and Benchmark:} We introduce \textbf{MMEVerse}, a large-scale and unified multimodal corpus for instruction-tuned emotion understanding. It consolidates 12 publicly available datasets into a standardized format, comprising over 130K annotated clips for training and 36K for testing across 18 evaluation benchmarks, thereby providing a consistent and extensible resource for reproducible evaluation and comparative analysis in affective intelligence research.

    \item \textbf{\pxj{Emotion-LLaMAv2 Architecture:}} We develop \pxj{an end-to-end} multimodal \pxj{emotion} large language model that eliminates explicit facial detection, and incorporates a \pxj{Conv Attention based fusion mechanism with spatiotemporal visual tokens and acoustic tokens, enabling simultaneous local and global multimodal feature interactions to capture complex emotional clues.}
    % \textcolor{red}{Conv Attention based fusion mechanism with temporal and acoustic token modeling}, and enables \textcolor{red}{fine grained multimodal representation learning} through cross modal attention.
    
    \item \pxj{\textbf{Perception-to-Cognition Training Framework}: We develop a curriculum learning strategy for Emotion-LLaMAv2 training, which aligns with human emotional development. It initially emphasizes fundamental emotion recognition, enabling the model to establish a solid foundation in identifying emotional signals. Subsequently, it integrates emotion reasoning training, incorporating multimodal emotional cues and contextual understanding to enhance the model's capability for nuanced emotional comprehension. }

    \item \textbf{Comprehensive Evaluation and Analysis:} We conduct extensive experiments and ablation studies across emotion recognition, emotion reasoning, and cross dataset generalization tasks. The results demonstrate consistent performance improvements over prior approaches and provide a reproducible experimental basis for future research on multimodal emotion understanding.
\end{itemize}

The remainder of this article is organized as follows. 
Section~\ref{sec:related} reviews prior research on multimodal large language models, instruction tuning, and emotion understanding. 
Section~\ref{sec:data} details the construction of the MMEVerse dataset and benchmark. 
Section~\ref{sec:model} describes the architecture and training scheme of Emotion LLaMAv2. 
Section~\ref{sec:experiments} presents the experimental setup, results, and analyses. 
Finally, Section~\ref{sec:conclusion} concludes the paper and outlines future directions.

\section{Related Work}
\label{sec:related}
To contextualize our contributions, this section reviews three relevant research areas: (1)~multimodal emotion understanding, (2)~multimodal large language models, and (3)~instruction tuning for cross-modal learning.

\subsection{Multimodal Emotion Understanding}
Multimodal emotion understanding aims to integrate visual, auditory, and textual cues to identify human affective states (emotion recognition) and to infer their underlying causes (emotion reasoning)~\cite{geetha2024multimodal, liu2024affective}. Early research in this area primarily adopted discriminative modeling paradigms under controlled laboratory settings, as exemplified by datasets such as IEMOCAP~\cite{busso2008iemocap}. Subsequent efforts extended these approaches to more diverse and unconstrained environments, including movies, television series, and in-the-wild scenarios~\cite{dhall2016emotiw, poria2018meld, jiang2020dfew, liu2022mafw}. Conventional methodologies typically extracted unimodal representations, such as CNN-based facial features~\cite{wang2020suppressing, li2020suppressing} or acoustic features learned by HuBERT~\cite{hsu2021hubert} and Wav2Vec~2.0~\cite{baevski2020wav2vec}, and subsequently integrated them through feature-level concatenation or attention-based fusion mechanisms~\cite{zadeh2017tensor, zadeh2018multimodal, zhang2023learning, li2023decoupled, wang2024incomplete}. While effective for classification-oriented tasks, these architectures primarily model low-level feature correlations and remain limited in their ability to support high-level semantic reasoning required to explain emotional inferences.

More recently, a paradigm shift has emerged with the development of Emergent Emotional Intelligence (EI), which emphasizes holistic, semantically grounded emotion understanding beyond surface-level classification. Initial benchmarks such as EmoBench~\cite{sabour2024emobench} and EmotionBench~\cite{huang2023emotionbench} evaluated this capability in text-based large language models. Building on this foundation, Emotion-LLaMA~\cite{cheng2024emotion} was the first to introduce an instruction-tuned multimodal large language model for emotion reasoning. This line of research was further advanced by subsequent work: AffectGPT~\cite{lian2025affectgpt} employed a Q-Former-based architecture to align multimodal features, while HumanOmni~\cite{zhao2025humanomni} and Omni-Emotion~\cite{yang2025omni} scaled training with large video corpora to capture omni-modal interactions. More recently, Audio-Reasoner~\cite{xie2025audio} and R1-Omni~\cite{zhao2025r1omni} enhanced reasoning performance through audio-centric modeling and reinforcement learning strategies.
Despite these methodological advances, existing evaluation benchmarks~\cite{liu2024emotion, hu2025emobenchm, zhang2025mmeemotion} remain fragmented in coverage and inconsistent in evaluation protocols, limiting their ability to support standardized training and rigorous comparison of multimodal emotion reasoning models. To address this gap, our work introduces \textbf{MMEVerse}, a unified large-scale corpus that consolidates twelve datasets into a coherent multimodal framework, enabling robust end-to-end training and standardized evaluation for multimodal emotion understanding.

\subsection{Instruction Tuning for LLM-based Emotion Understanding}
Instruction tuning~\cite{ouyang2022instruct-tuning, wei2021finetuned} has emerged as a paradigm for aligning Large Language Models (LLM) with human intent and requirements. Foundational work in natural language processing, including InstructGPT~\cite{ouyang2022instruct-tuning}, FLAN~\cite{chung2022scaling}, Self-Instruct~\cite{wang2022self_instruct}, and OPT-IML~\cite{iyer2022opt-iml}, demonstrated that learning from diverse instruction–response pairs substantially improves zero-shot and few-shot generalization. Building on this paradigm, multimodal extensions such as LLaVA~\cite{liu2024visual} and LLaVA-NeXT~\cite{liu2024llava} successfully adapted instruction tuning to vision–language tasks, enabling models to follow natural language instructions grounded in visual content.
While supervised fine-tuning (SFT) provides a basic form of alignment, recent approaches have incorporated preference optimization and reinforcement learning (RL) strategies to enhance reasoning capability. Representative examples include DeepSeek-R1~\cite{guo2025deepseekr1} and Visual-RFT~\cite{liu2025visualrft}, which leverage reward-driven optimization to encourage structured and logically consistent responses. In the affective domain, R1-Omni~\cite{zhao2025r1omni} extends this idea by applying reinforcement learning with verified feedback (RLVR) to enforce logical coherence in emotional reasoning.
Despite these advances, instruction tuning for multimodal emotion understanding remains underexplored. Earlier efforts such as EmoVIT~\cite{xie2024emovit} and the initial MERR dataset~\cite{cheng2024emotion} primarily relied on caption-based instruction generation with limited scale, restricted modality coverage, and weak support for reasoning-oriented supervision. These limitations constrain the ability of models to generalize across diverse emotional contexts and to develop cognitively grounded affective reasoning. In contrast, our \textbf{MMEVerse} introduces a large-scale, unified instruction-tuning benchmark that employs a multi-agent pipeline to generate recognition- and reasoning-oriented instructions aligned with multimodal emotional evidence. This design provides a foundation for training the next generation of multimodal emotional intelligence systems.

\begin{table*}[t]
\setlength{\abovecaptionskip}{0pt}
\setlength{\belowcaptionskip}{1pt}
\centering
\caption{\textbf{Summary of datasets integrated into our MMEVerse}. Original training splits are merged to form \textbf{MMEVerse-Train}, while validation and test splits constitute \textbf{MMEVerse-Bench}. Each sample is carefully re-annotated with detailed multimodal emotional descriptions, ensuring consistency across datasets and enabling unified benchmarking.}
\resizebox{\linewidth}{!}{
\renewcommand\arraystretch{0.95}
\tiny
\begin{tabular}{l||rlccl}
\thickhline
\rowcolor[RGB]{230, 230, 230}\textbf{Dataset} & \textbf{\# Samples} & \textbf{Annotation Type} & \textbf{\# Labels}& \textbf{Description}& \textbf{Source} \\
\hline\hline
\rowcolor[RGB]{245, 245, 245}MER2023~\cite{lian2023mer}    & 3784        & basic emotion & 6 & \textcolor{myred}{\xmark}  & movies and TV series \\
MELD~\cite{poria2018meld}       & 13798      & basic emotion, sentiment & 7, 3 & \textcolor{myred}{\xmark}  & ``Friends'' TV series \\
\rowcolor[RGB]{245, 245, 245}IEMOCAP~\cite{busso2008iemocap}     & 10039        & basic emotion, dimensiona (VAD) & 7 & \textcolor{myred}{\xmark}  & in-the-lab actor performance \\
CAER~\cite{lee2019context}      & 13175        & basic emotion & 7 & \textcolor{myred}{\xmark}  &  movies \\
\rowcolor[RGB]{245, 245, 245}E3~\cite{feng20243}         & 16320        & basic emotion, sentiment & 8, 3 & \mygreencheck & egocentric videos \\
DFEW~\cite{jiang2020dfew}       & 11697        & basic emotion & 7 & \textcolor{myred}{\xmark}  & movies \\
\rowcolor[RGB]{245, 245, 245}MAFW~\cite{liu2022mafw}       & 10045        & single emotion, multi-label emotions & 11, 399 & \mygreencheck & movies \\
MC-EIU~\cite{liu2024emotion}     & 10830        & basic emotion, intention & 7, 9 & \textcolor{myred}{\xmark}  & TV series (3 en \& 4 zh) \\
\rowcolor[RGB]{245, 245, 245}CMU-MOSI~\cite{zadeh2017tensor}   & 2199        & sentiment & [-3, 3] &\textcolor{myred}{\xmark}   & videos in YouTube \\
CMU-MOSEI~\cite{zadeh2018multimodal}  & 22856       & sentiment &[-3, 3] & \textcolor{myred}{\xmark}  & videos in YouTube \\
\rowcolor[RGB]{245, 245, 245}CH-SIMS v2~\cite{liu2022make}  & 4403      & sentiment & 7 & \textcolor{myred}{\xmark} & diverse Mandarin videos \\
BOLD~\cite{luo2020arbee}       & 10072        & single emotion, dimension (VAD)& 26 & \textcolor{myred}{\xmark}  & movies in YouTube \\
\thickhline
\end{tabular}}
\vspace{-0.2cm}
\label{tab:dataMMEVerse}
\end{table*}

% \section{MMEVerse: Multimodal Emotion Corpus}
% \label{sec:data}
% Robust multimodal emotion understanding requires datasets that offer tightly synchronized video, audio, and textual streams together with reliable affective supervision. Existing corpora, however, remain limited in several respects: modalities are often only partially aligned, annotation protocols vary widely across datasets, and both emotional taxonomies and descriptive depth tend to be restricted. Such inconsistencies hinder the development of multimodal large language models (MLLMs) capable of strong generalization and coherent cross-modal reasoning. To overcome these limitations, we present \textbf{MMEVerse}, a unified multimodal emotion corpus that aggregates 12 widely used datasets and re-annotates all samples through a scalable, fine-grained multimodal framework. By combining broad coverage with consistent, high-quality descriptions, MMEVerse establishes a reliable foundation for advancing both recognition and reasoning in affective computing.

\section{MMEVerse: Multimodal Emotion Corpus}
\label{sec:data}
Robust multimodal emotion understanding depends on datasets that provide tightly synchronized video, audio, and textual streams together with reliable and semantically consistent affective supervision. However, existing multimodal emotion corpora remain limited in several critical aspects. Modalities are often only partially aligned, annotation protocols and emotional taxonomies vary substantially across datasets, and most resources focus on categorical labels while lacking descriptive signals that support deeper cross-modal reasoning. These inconsistencies significantly hinder the development of multimodal large language models (MLLMs) that require both large-scale supervision and coherent multimodal semantics to achieve strong generalization and interpretable emotional inference.

To address these limitations, we introduce \textbf{MMEVerse}, a unified multimodal emotion corpus that aggregates 12 widely used datasets and re-annotates all samples through a scalable and fine-grained multimodal annotation framework. Rather than simply merging heterogeneous sources, MMEVerse enforces strict tri-modal synchronization, consistent emotional supervision, and uniform descriptive structure across diverse domains. By combining broad coverage with high-quality and systematically generated multimodal descriptions, MMEVerse establishes a principled foundation for advancing both emotion recognition and multimodal emotion reasoning in affective computing.

% \subsection{Data Curation}
% \label{sec:data_curation}
% We begin by systematically surveying multimodal emotion datasets employed in affective computing, human–computer interaction, and multimodal sentiment analysis. Only datasets offering \emph{fully synchronized} video, audio, and textual modalities are retained to ensure suitability for models that jointly reason over facial dynamics, prosodic clues, and linguistic content. Based on this criterion, we select 12 representative datasets spanning controlled laboratory recordings, scripted cinematic content, television series, conversational interactions, and in-the-wild social media videos. This diversity enables \textbf{MMEVerse} to capture cultural, linguistic, and contextual variation in emotional expression. Table~\ref{tab:dataMMEVerse} summarizes the datasets and their annotation formats. To maintain semantic consistency across sources, we preserve only the original categorical emotion labels and discard prior descriptive annotations. Each clip is re-annotated using a unified fine-grained pipeline (Section~\ref{sec:annotation}). After alignment and standardization, \textbf{MMEVerse} comprises 129,128 tri-modal video clips, establishing it as the largest corpus with synchronized multimodal inputs and uniform emotional supervision.

\subsection{Data Curation}
\label{sec:data_curation}
We begin by systematically surveying existing multimodal emotion datasets used in affective computing, human–computer interaction, and multimodal sentiment analysis. To ensure suitability for models that perform joint multimodal reasoning, we retain only datasets that provide fully synchronized video, audio, and textual streams, enabling coherent analysis of facial dynamics, prosodic cues, and linguistic content. Based on this criterion, we select 12 representative datasets spanning controlled laboratory recordings, scripted cinematic content, television series, conversational interactions, and in-the-wild social media videos. This diversity allows \textbf{MMEVerse} to capture broad cultural, linguistic, and contextual variation in emotional expression.
Table~\ref{tab:dataMMEVerse} summarizes the selected datasets and their original annotation formats. To maintain semantic consistency across heterogeneous sources, we preserve only the original categorical emotion labels and discard prior descriptive annotations. Each clip is then re-annotated using a unified fine-grained pipeline described in Section~\ref{sec:annotation}. After alignment and standardization, \textbf{MMEVerse} comprises 129{,}128 tri-modal video clips, establishing it as the largest corpus with strictly synchronized multimodal inputs and uniform emotional supervision.

\subsection{Annotation Pipeline}
\label{sec:annotation}

High-quality multimodal supervision is essential for enabling MLLMs to perform coherent and interpretable emotional reasoning. To support annotation at scale while maintaining semantic consistency across heterogeneous data sources, we design a unified multimodal annotation pipeline that systematically re-annotates each sample through structured visual, acoustic, and linguistic analyses. For each video, Facial Action Units (AUs) are first extracted from all frames using OpenFace, and the frame with the highest aggregated AU intensity score $S_{\text{au}_i}$ is selected as the peak emotional frame, which serves as the anchor for subsequent visual expression analysis. This peak frame is then processed by two complementary vision models: OpenFace~\cite{baltruvsaitis2016openface} produces AU-based facial expression descriptors characterizing fine-grained muscle movements, while Qwen2.5-VL-72B~\cite{bai2025qwen2} extracts high-level contextual information regarding scene layout, objects, and interactions. These outputs form the visual expression description ($C_{ved}$) and the visual objective description ($C_{vod}$), respectively. In parallel, the audio stream is analyzed using Qwen2-Audio~\cite{chu2024qwen2} or Audio-Reasoner~\cite{xie2025audio} to extract prosodic and paralinguistic cues such as pitch variation, speaking rate, intensity, pauses, and hesitations, yielding the audio tone description ($C_{atd}$), which complements visual cues particularly when facial signals are subtle or ambiguous. Lexical subtitles ($C_{ls}$) further provide semantic and pragmatic context that grounds emotional expressions in spoken content. All unimodal clues, including $C_{ved}$, $C_{vod}$, $C_{atd}$, and $C_{ls}$, are then integrated to form a preliminary multimodal emotional description. This description is subsequently refined using GPT-4o~\cite{hurst2024gpt}, which synthesizes unimodal evidence into a coherent and well-structured multimodal emotional description ($C_{md}$), resolving potential cross-modal conflicts and enforcing consistent phrasing and reasoning quality. The consolidated description, together with the original categorical emotion label, constitutes the final annotation for each sample. To assess annotation reliability, a subset of samples is further evaluated through a human user study~\cite{cheng2024shield}, providing quantitative evidence of annotation quality and consistency. The overall procedure is summarized in Algorithm~\ref{alg:multimodal_emotion_annotation} and visualized in the supplement.

\begin{algorithm}[t]
\caption{Multimodal Emotion Annotation Procedure}
\label{alg:multimodal_emotion_annotation}
% \hrule % 添加横线
\begin{algorithmic}[1]
\Require Video frames $V = {f_1, f_2, \ldots, f_n}$, Audio stream $A$, Lexical Subtitles.
\Ensure Annotated video with comprehensive emotional descriptors for the peak emotional expression frame.
\State Initialize $I_{\text{peak}} \gets 0$;
\State Initialize $Frame_{\text{peak}} \gets \emptyset$;
\For{each frame $f_k$ in $V$}
\State Detect AUs and compute $I_{\text{AU}} \gets \sum_{i} S^k_{au_i}$;
\If{$I_{\text{AU}} > I_{\text{peak}}$}
\State $I_{\text{peak}} \gets I_{\text{AU}}$;
\State $Frame_{\text{peak}} \gets f_k$;
\EndIf
\EndFor
\State Analyze $Frame_{\text{peak}}$ with OpenFace to obtain $C_{ved}$;
\State Analyze $Frame_{\text{peak}}$ with Qwen2.5-VL-72B to obtain $C_{vod}$;
\State Analyze audio $A$ with Qwen2-Audio or Audio-Reasoner to obtain $C_{atd}$;
\State Integrate $C_{ls}$, $C_{ved}$, $C_{vod}$, and $C_{atd}$ to synthesize context;
\State Generate comprehensive description $C_{md}$ with GPT4-o;
\State Annotate $Frame_{\text{peak}}$ with $C_{md}$;
\State \Return Annotated $Frame_{\text{peak}}$.
\end{algorithmic}
\end{algorithm}

\begin{table}[t]
\setlength{\abovecaptionskip}{0pt}
\setlength{\belowcaptionskip}{0pt}
\centering
\caption{Dataset statistics for \textbf{MMEVerse-Bench}.}
\resizebox{\linewidth}{!}{
\renewcommand\arraystretch{1.15}
\scriptsize
\begin{tabular}{clcr}
\thickhline
\rowcolor[RGB]{230, 230, 230}\textbf{Unibench} & \textbf{Source} & \textbf{Chosen Set} & \textbf{\# Samples} \\
\hline\hline
\multicolumn{4}{c}{\textit{\textbf{Basic Emotion}}} \\
\hline
\rowcolor[RGB]{245, 245, 245}\mygreencheck &  MER2023~\cite{lian2023mer}            & MER-MULTI & 411 \\
\mygreencheck & MER2024~\cite{lian2024mer}
           & MER-SEMI & 1169 \\
\rowcolor[RGB]{245, 245, 245}\mygreencheck & MELD-e~\cite{poria2018meld}            & Test & 2610 \\
\mygreencheck & IEMOCAP~\cite{busso2008iemocap}             & Session5 & 1241 \\
\rowcolor[RGB]{245, 245, 245}\textcolor{myred}{\xmark} & CAER~\cite{lee2019context}                 & Test,Val & 3953 \\
\textcolor{myred}{\xmark} & E3~\cite{feng20243}                   & Test,Val & 3288 \\
\rowcolor[RGB]{245, 245, 245}\textcolor{myred}{\xmark} & DFEW~\cite{jiang2020dfew}                 & Set1-Test & 2341 \\
\textcolor{myred}{\xmark} & MAFW~\cite{liu2022mafw}                 & Test & 1839 \\
\rowcolor[RGB]{245, 245, 245}\textcolor{myred}{\xmark} & MC-EIU~\cite{liu2024emotion}               & Test,Val & 3249 \\
\hline
\multicolumn{4}{c}{\textit{\textbf{Sentiment}}} \\
\hline
\rowcolor[RGB]{245, 245, 245}\mygreencheck  & CMU-MOSI~\cite{zadeh2017tensor}             & Test & 686 \\
\mygreencheck  & CMU-MOSEI~\cite{zadeh2018multimodal}            & Test & 4659 \\
\rowcolor[RGB]{245, 245, 245}\mygreencheck  & CH-SIMS~\cite{yu2020ch}                 & Test & 457 \\
\mygreencheck  & CH-SIMS v2~\cite{liu2022make}         & Test & 1034 \\
\rowcolor[RGB]{245, 245, 245}\textcolor{myred}{\xmark} & MELD-s~\cite{poria2018meld}               & Test, Dev & 3719 \\
\hline
\multicolumn{4}{c}{\textit{\textbf{Multi-Label}}} \\
\hline
\rowcolor[RGB]{245, 245, 245}\textcolor{myred}{\xmark}   & MAFW-m~\cite{liu2022mafw}               & Test & 580 \\
\textcolor{myred}{\xmark}   & BOLD~\cite{luo2020arbee}                 & Val & 1116 \\
\hline
\multicolumn{4}{c}{\textit{\textbf{OV-Emotion}}} \\
\hline
\rowcolor[RGB]{245, 245, 245}\mygreencheck & OV-MERD+~\cite{lian2025affectgpt}            & ALL & 532 \\
\hline
\multicolumn{4}{c}{\textit{\textbf{Intention}}} \\
\hline
\rowcolor[RGB]{245, 245, 245}\textcolor{myred}{\xmark}  & MC-EIU-i~\cite{liu2024emotion}             & Test, Val & 3249 \\
\thickhline
\end{tabular}}
\vspace{-0.4cm}
\label{tab:mmeverse-bench}
\end{table}

% \subsection{Benchmark Construction}
% \label{sec:benchmark}
% All re-annotated samples are organized into \textbf{MMEVerse-Train} and \textbf{MMEVerse-Bench}, retaining the original training, validation, and test partitions of each dataset to ensure strict separation and avoid information leakage. Because emotional taxonomies vary considerably across sources, MMEVerse-Bench adopts an evaluation protocol that preserves the native label space of each dataset, thereby preventing incompatible category alignment and more faithfully capturing the diversity of real-world affective constructs. The benchmark comprises 36{,}133 clips spanning basic and complex emotions, multi-label affect, sentiment polarity, valence–arousal dimensions, and affective intents. It fully incorporates MER-UniBench~\cite{lian2025affectgpt} while extending it with nine additional evaluation sets that broaden emotional and contextual coverage. 
% Comprehensive summary statistics for all benchmark components are provided in Table~\ref{tab:mmeverse-bench}.

\subsection{Benchmark Construction}
\label{sec:benchmark}
All re-annotated samples are organized into two disjoint subsets, \textbf{MMEVerse-Train} and \textbf{MMEVerse-Bench}, by strictly retaining the original training, validation, and test partitions provided by each source dataset. This design ensures a clear separation between data used for model optimization and data reserved exclusively for evaluation, thereby avoiding information leakage and enabling fair and reproducible benchmarking across different learning paradigms.
Because emotional taxonomies and annotation schemes vary substantially across existing corpora, MMEVerse-Bench adopts an evaluation protocol that preserves the native label space of each dataset rather than enforcing a unified category mapping. This choice prevents incompatible or lossy alignment between heterogeneous emotion definitions and allows evaluation results to more faithfully reflect the diversity and subjectivity of real-world affective constructs. As a result, models are assessed within the semantic scope originally defined by each dataset, ensuring that performance comparisons remain meaningful and interpretable.
The resulting benchmark comprises 36{,}133 clips spanning a wide range of affective settings, including basic and complex emotions, multi-label affect, sentiment polarity, valence–arousal dimensions, and affective intents. MMEVerse-Bench fully incorporates MER-UniBench~\cite{lian2025affectgpt} as a core component, while further extending it with nine additional evaluation sets that substantially broaden emotional coverage, contextual diversity, and modality combinations. This expanded benchmark enables comprehensive assessment of multimodal emotion recognition and reasoning models under varied semantic, cultural, and situational conditions.
Comprehensive summary statistics and detailed composition of all benchmarks are provided in Table~\ref{tab:mmeverse-bench}.

\subsection{Dataset Comparison}
\label{sec:dataset_comparison}
To contextualize the contribution of MMEVerse within existing multimodal emotion resources, we compare it with representative corpora commonly used in multimodal emotion analysis, presenting in Table \ref{tab:datasets_compare}. Prior datasets such as IEMOCAP~\cite{busso2008iemocap}, MELD~\cite{poria2018meld}, DFEW~\cite{jiang2020dfew}, and MER2023~\cite{lian2023mer} primarily provide categorical labels without expressive descriptions that integrate scene context, facial dynamics, and acoustic prosody, whereas EmoSet~\cite{yang2023emoset} and EmoVIT~\cite{xie2024emovit} offer visual descriptions but omit audio clues and cross-modal reasoning signals. EMER~\cite{lian2023explainable} includes high-quality multimodal descriptions but remains too small to effectively support the training of modern multimodal LLMs. Intermediate resources including MERR-Coarse~\cite{cheng2024emotion}, MERR-Fine~\cite{cheng2024emotion}, MER-Caption~\cite{lian2025affectgpt}, and MER-Caption+~\cite{lian2025affectgpt} aim to enhance emotional supervision through model-generated annotations with limited human verification, yet their scale and annotation consistency are insufficient for robust generalization across heterogeneous sources. In contrast, MMEVerse offers 129{,}128 tri-modally aligned clips with fine-grained and consistently generated multimodal descriptions and broad affective coverage encompassing basic and complex emotions, multi-label affect, sentiment polarity, and valence and arousal dimensions.

\begin{table*}[ht]
\setlength{\abovecaptionskip}{0pt}
\setlength{\belowcaptionskip}{2pt}
\scriptsize
\centering
\caption{\textbf{Comparison of representative emotion datasets across key annotation dimensions.}
MMEVerse is the only corpus that provides large-scale data with unified multimodal descriptions, 
covering audio, visual objective clues, visual expression clues (AUs), label-guided descriptions, 
model-based descriptions, and categorical emotion labels. }
% \vspace{4mm}
\resizebox{\linewidth}{!}{
\renewcommand\arraystretch{1.1}
\tiny
\begin{tabular}{l||cccccccc}
\thickhline
\rowcolor[RGB]{230, 230, 230}& \begin{tabular}[c]{@{}c@{}}\textbf{Sufficient} \\ \textbf{Quantity} \end{tabular} 
& \begin{tabular}[c]{@{}c@{}} \textbf{Audio}     \\ \textbf{Description}\end{tabular} 
& \begin{tabular}[c]{@{}c@{}} \textbf{Visual Objective} \\ \textbf{Description}\end{tabular} 
& \begin{tabular}[c]{@{}c@{}} \textbf{Visual Expression}\\ \textbf{Description}\end{tabular} 
& \begin{tabular}[c]{@{}c@{}} \textbf{Label-guided  }\\ \textbf{Description} \end{tabular} 
& \begin{tabular}[c]{@{}c@{}} \textbf{Model-based}  \\ \textbf{Description} \end{tabular} 
& \begin{tabular}[c]{@{}c@{}} \textbf{Classification} \\ \textbf{Label} \end{tabular} 
& \begin{tabular}[c]{@{}c@{}}\textbf{Multimodal}\\ \textbf{Description}\end{tabular} 
\\
\hline\hline
\rowcolor[RGB]{245, 245, 245}EmoSet~\cite{yang2023emoset} 
& \mygreencheck & \textcolor{myred}{\xmark} & \mygreencheck & \textcolor{amber}{\halfcheckmark} & \textcolor{myred}{\xmark} & \mygreencheck & \textcolor{amber}{\halfcheckmark} & \textcolor{myred}{\xmark}  \\[0.5ex]
 EmoVIT ~\cite{xie2024emovit} 
& \mygreencheck & \textcolor{myred}{\xmark} & \mygreencheck & \textcolor{amber}{\halfcheckmark} & \textcolor{myred}{\xmark} & \mygreencheck & \textcolor{amber}{\halfcheckmark} & \textcolor{amber}{\halfcheckmark} \\[0.5ex]
\rowcolor[RGB]{245, 245, 245}DFEW~\cite{jiang2020dfew} 
& \mygreencheck & \textcolor{myred}{\xmark} & \textcolor{myred}{\xmark} & \textcolor{myred}{\xmark} & \textcolor{myred}{\xmark} & \textcolor{myred}{\xmark} & \mygreencheck & \textcolor{myred}{\xmark} \\[0.5ex]
MER2023 ~\cite{lian2023mer}
& \mygreencheck & \textcolor{myred}{\xmark} & \textcolor{myred}{\xmark} & \textcolor{myred}{\xmark} & \textcolor{myred}{\xmark} & \textcolor{myred}{\xmark} & \mygreencheck & \textcolor{myred}{\xmark} \\[0.5ex]
\rowcolor[RGB]{245, 245, 245}MAFW~\cite{liu2022mafw} 
& \mygreencheck & \textcolor{myred}{\xmark} & \mygreencheck & \mygreencheck & \textcolor{myred}{\xmark} & \textcolor{myred}{\xmark} & \mygreencheck & \textcolor{myred}{\xmark} \\[0.5ex]
EMER~\cite{lian2023explainable} 
& \textcolor{myred}{\xmark} & \mygreencheck & \mygreencheck & \mygreencheck& \textcolor{myred}{\xmark}&   \mygreencheck & \mygreencheck & \mygreencheck \\
\hline
\rowcolor[RGB]{245, 245, 245}MERR-Coarse~\cite{cheng2024emotion}   & \mygreencheck & \mygreencheck & \mygreencheck & \mygreencheck & \textcolor{myred}{\xmark} & \mygreencheck&  \mygreencheck & \mygreencheck  \\
MERR-Fine~\cite{cheng2024emotion}  & \mygreencheck & \mygreencheck & \mygreencheck & \mygreencheck & \textcolor{myred}{\xmark} & \mygreencheck & \mygreencheck & \mygreencheck  \\
\rowcolor[RGB]{245, 245, 245}MER-Caption~\cite{lian2025affectgpt}  & \mygreencheck & \mygreencheck & \mygreencheck & \mygreencheck & \textcolor{myred}{\xmark} & \mygreencheck & \mygreencheck & \mygreencheck  \\
MER-Caption+~\cite{lian2025affectgpt}  & \mygreencheck & \mygreencheck & \mygreencheck & \mygreencheck & \textcolor{myred}{\xmark} & \mygreencheck & \mygreencheck & \mygreencheck  \\
\rowcolor[RGB]{245, 245, 245}MMEVerse & \mygreencheck & \mygreencheck & \mygreencheck & \mygreencheck & \mygreencheck & \mygreencheck & \mygreencheck & \mygreencheck  \\
\thickhline
\end{tabular}}
\label{tab:datasets_compare}
\end{table*}

% \section{Methodology}
% \label{sec:model}
% \begin{figure*}[h]
% \centering
% \includegraphics[width=\linewidth]{images/pipeline3.png}
% \caption{\textbf{Overall Architecture of Emotion-LLaMAv2.} Emotion-LLaMAv2 integrates audio, visual, and textual inputs for multimodal emotion recognition and reasoning. The framework comprises a multiview encoder for extracting unimodal embeddings, a Conv-Attention module for pre-fusing audio and visual signals, a model adapter to align modality-specific embeddings with LLMs, and a LoRA-tuned LLM for downstream reasoning.}
% \label{fig:framework}
% \vspace{-0.3cm}
% \end{figure*}

\section{Methodology}
\label{sec:model}
\begin{figure*}[t]
\setlength{\abovecaptionskip}{2pt}
\setlength{\belowcaptionskip}{2pt}
\centering
\includegraphics[width=\linewidth]{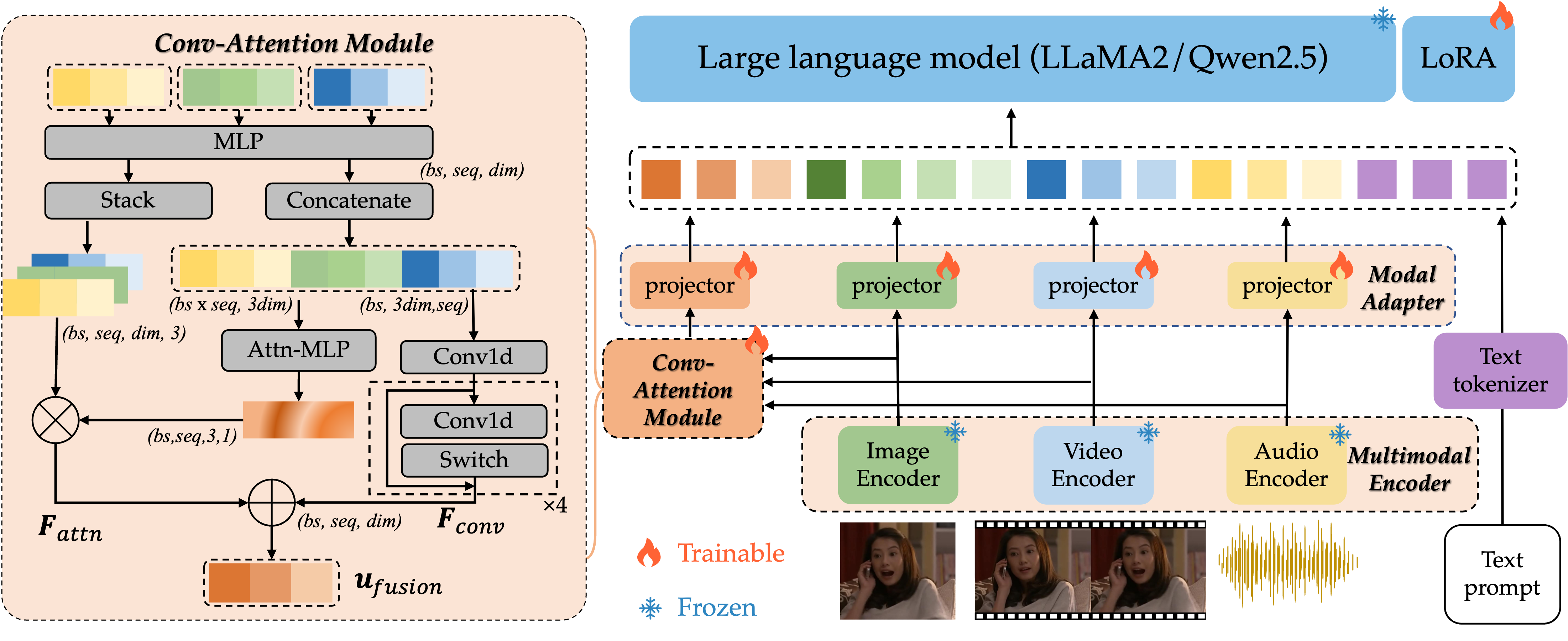}
\caption{\textbf{Overall architecture of Emotion-LLaMAv2.} 
The framework processes audio, visual, and textual inputs through a structured pipeline for multimodal emotion recognition and reasoning. Modality-specific encoders extract unimodal representations, which are integrated via a Conv-Attention pre-fusion module to model emotion-aware cross-modal interactions. The fused features are then aligned to the language model embedding space through a modal adapter, enabling a LoRA-tuned large language model to perform joint emotion recognition and multimodal reasoning.}
\label{fig:framework}
\vspace{-0.3cm}
\end{figure*}

This section presents the methodological details of Emotion-LLaMAv2. We first introduce the overall architectural design and its underlying principles, followed by a detailed description of each component, including multimodal encoding, pre-fusion with Conv-Attention, representation alignment via modal adapters, and the two-stage instruction-tuning strategy for emotion recognition and reasoning.

\subsection{Overview}
\label{sec:overview}

Emotion-LLaMAv2 presents an end-to-end framework for multimodal emotion recognition and reasoning, as illustrated in Figure~\ref{fig:framework}. Unlike most existing MLLMs or emotion-oriented systems that primarily rely on modality-specific feature extraction followed by shallow projection into a language model, Emotion-LLaMAv2 is explicitly designed to support structured affective perception, multimodal fusion, and reasoning within a unified architecture. The framework decomposes multimodal emotion understanding into four tightly coupled components: a multimodal encoder for perception, a Conv-Attention pre-fusion module for emotion-aware cross-modal interaction, a modal adapter for representation alignment, and a LoRA-tuned large language model for joint emotion recognition and reasoning.

This modular design reflects the hierarchical nature of human affect understanding. Low-level perceptual signals are first extracted independently from each modality, then selectively fused to capture emotion-relevant interactions, aligned into a shared representational space, and finally interpreted through language-based reasoning. By explicitly structuring the processing pipeline in this manner, Emotion-LLaMAv2 avoids overloading the language model with raw multimodal fusion and instead enables more stable, interpretable, and controllable affective inference.

Formally, let $\mathcal{E}^{a}$ denote the audio encoder, $\mathcal{E}^{v}$ the visual encoder, $\mathcal{F}^{av}$ the Conv-Attention pre-fusion module, and $\mathcal{\phi}$ the multimodal large language model. Given an input tuple
\[
P = \langle \text{Audio}, \text{Video}, \text{Prompt} \rangle,
\]
the output of Emotion-LLaMAv2 is defined as:
\begin{equation}
    \setlength\abovedisplayskip{3pt}
    \setlength\belowdisplayskip{3pt}
\begin{aligned}
\hat{O} &= \Psi(\mathcal{\phi}, \mathcal{E}, \Omega, \mathcal{F}, P) \\
&= \mathcal{\phi}\big(
    \mathcal{E}^{a}(\text{A}),
    \mathcal{E}^{v}(\Omega(\text{V})),
    \mathcal{F}^{av}(\text{A}, \Omega(\text{V})),
    \mathcal{E}^{t}(\text{Prompt})
\big),
\label{eq:emotion_llama}
\end{aligned}
\end{equation}
where $\Omega$ denotes the vision pre-processing operator and $\mathcal{E}$ represents the collection of modality-specific encoders. The output $\hat{O}$ corresponds to the formatted textual response generated by the language model, which may represent either an emotion category label or a reasoning explanation, depending on the task instruction. The input video $V$ is pre-processed into both a representative frame and a frame sequence, as detailed in Section~\ref{sec:mmEncoder}, enabling the model to capture both static and dynamic emotional cues.

\textbf{Multimodal Prompt Template}. 
To support emotion recognition and reasoning within a unified language modeling framework, we design a structured multimodal prompt template that integrates modality-aware feature tokens with natural language instructions. The prompt explicitly exposes fused multimodal representations to the language model while preserving their semantic roles, guiding the model to associate visual and auditory evidence with affective interpretations. The general structure of the prompt is defined as:
\begin{quote}
\texttt{[Vid]} $\mathtt{<FusionFeature>}$ $\mathtt{<ImageFeature>}$ $\mathtt{<VideoFeature>}$ $\mathtt{<AudioFeature>}$ \texttt{[Vid]} \\
The person in the video says: $\mathtt{<Text>}$ $\mathtt{<Task Identifier>}$ Prompt.
\end{quote}

This structured formulation enables the language model to attend selectively to different modality tokens while conditioning its output on task-specific instructions. As a result, Emotion-LLaMAv2 can flexibly support both categorical emotion prediction and fine-grained multimodal emotion reasoning within a single autoregressive framework.
With this architectural overview, we \pxj{detail} each component of Emotion-LLaMAv2 in \pxj{the following}. 
% Section~\ref{sec:mmEncoder} introduces the multimodal encoders and tokenization strategy for extracting complementary audio and visual representations. Section~\ref{sec:conv-attention} presents the Conv-Attention pre-fusion module for modeling emotion-aware cross-modal interactions. Section~\ref{sec:modal-adapter} details the modal adapter that aligns heterogeneous multimodal features with the language model embedding space. Finally, Section~\ref{sec:training} outlines the two-stage instruction-tuning strategy that integrates emotion recognition and emotion reasoning into a unified training framework.

\subsection{Multimodal Encoder and Tokenization}
\label{sec:mmEncoder}

\textbf{Audio and Visual Encoders}. 
Capturing emotional states from multimodal signals requires representations that are both expressive and complementary across modalities. To this end, Emotion-LLaMAv2 employs an audio encoder $\mathcal{E}^{a}$ together with a multiview visual encoder $\mathcal{E}^{v}$, enabling the model to jointly model prosodic, facial, and contextual cues that are critical for affective understanding.
For the audio modality, we adopt a pre-trained audio encoder $\mathcal{E}^{a}$, instantiated as either HuBERT~\cite{hsu2021hubert} or Whisper~\cite{radford2023robust}, both of which have demonstrated strong robustness and generalization in speech-related tasks. Given an input audio signal, the encoder extracts a high-level auditory representation $u^{a}$ that captures prosodic patterns such as intonation, rhythm, and energy variation, which are known to be highly informative for emotion recognition. Importantly, the use of pre-trained audio encoders allows Emotion-LLaMAv2 to leverage large-scale speech representations while remaining agnostic to language and speaker identity.
For the visual modality, emotional expressions are inherently multi-scale and temporally dynamic. Relying on a single visual encoder is often insufficient to capture both instantaneous facial expressions and longer-term affective dynamics. Therefore, the vision preprocessor decomposes the input video into two complementary components, which are processed by a global visual encoder $\mathcal{E}^{v}_{glo}$ and a temporal visual encoder $\mathcal{E}^{v}_{temp}$, respectively. This dual-encoder design enables the extraction of multi-view visual emotional representations that jointly encode static appearance and temporal evolution.

The global visual encoder $\mathcal{E}^{v}_{glo}$ processes a representative frame sampled from the middle of the video sequence. This frame serves as a heuristic proxy for high emotional salience, as emotional expressions often peak near the central portion of short clips. We implement $\mathcal{E}^{v}_{glo}$ using a ViT-based architecture, EVA~\cite{fang2023eva}, initialized with official pre-trained weights. The resulting global visual feature,
% \[
$u^{v}_{glo} = \mathcal{E}^{v}_{glo}(\mathit{Frame})$
% \]
captures both facial expressions and broader scene context, allowing the model to reason about emotions in relation to environmental and situational cues.

The temporal visual encoder $\mathcal{E}^{v}_{temp}$ is designed to capture the dynamic aspects of emotional expression. Frames are uniformly sampled from the video sequence to preserve temporal coverage while maintaining computational efficiency. We explore two categories of temporal encoders. The first employs general video representation models, including VideoMAE~\cite{tong2022videomae} and VideoMAEv2~\cite{wang2023videomae}, which explicitly model temporal dependencies across frames. The second applies general image representation models, such as CLIP~\cite{radford2021learning}, independently to multiple frames, followed by pooling to obtain a compact temporal representation. Although these approaches differ in architectural assumptions, both operate on low-resolution spatial features and focus on capturing coarse temporal dynamics. The temporal encoder produces a dynamic emotional representation,
% \[
$u^{v}_{temp} = \mathcal{E}^{v}_{temp}(V)$,
% \]
which complements the global visual feature by encoding motion patterns, expression transitions, and short-term affective fluctuations.

In preliminary experiments with the original Emotion-LLaMA framework, an explicit face-sequence preprocessor was incorporated to extract cropped facial regions. However, we found that such preprocessing introduced additional complexity and potential error propagation, while providing limited performance gains when combined with strong visual backbones. As a result, Emotion-LLaMAv2 omits explicit face detection and operates directly on full-frame inputs, enabling a more streamlined and fully end-to-end training pipeline while allowing the model to implicitly attend to emotion-relevant regions.
It is worth noting that the optimal number of retained audio and visual tokens is not fixed and depends on factors such as sampling rate, temporal resolution, and spatial merging strategies. This trade-off directly affects both representational capacity and computational efficiency. We provide an extensive empirical analysis of this relationship in Section~\ref{sec:experiments}.

\textbf{Tokenization}. 
Textual inputs are processed using the LLaMA tokenizer, which adopts a byte-pair encoding (BPE) scheme implemented via SentencePiece~\cite{kudo2018sentencepiece}. This tokenizer is well suited for open-vocabulary settings and can robustly handle the diverse linguistic expressions commonly observed in emotion-related dialogue, including informal speech, interjections, and incomplete utterances. 
For multimodal emotional reasoning, we employ a modified autoregressive generation procedure in which tokens are iteratively selected based on their conditional likelihood given the joint multimodal context. This design allows the language model to dynamically integrate textual semantics with aligned audio and visual tokens, rather than treating language as an isolated modality. 

\subsection{Multimodal Pre-fusion with Conv-Attention Module}
\label{sec:conv-attention}
Effective fusion of multimodal information is a critical prerequisite for emotion understanding, as affective states are typically expressed through subtle and temporally evolving cues distributed across audio, visual, and linguistic channels. In our initial Emotion-LLaMA framework, multimodal fusion was implicitly delegated to the large language model itself. Although this design benefits from architectural simplicity, we observed that it was insufficient for handling the heterogeneity and temporal complexity of multimodal emotional signals, often leading to unstable optimization and ambiguous cross-modal associations during training.

Recent work such as AffectGPT~\cite{lian2025affectgpt} addresses this issue by introducing explicit cross-modal interaction modules outside the LLM, including Q-Former-based architectures and simple attention mechanisms. AffectGPT further demonstrates that average-pooled unimodal features combined with attention-weighted aggregation can outperform Q-Former in emotion-related tasks. While effective, these approaches rely on pooled representations that compress temporal structure and local dynamics, which are particularly important for modeling fine-grained emotional expressions such as micro-expressions, vocal prosody shifts, and short-lived affective transitions.

To overcome these limitations, we introduce a \textit{Conv-Attention} module for multimodal pre-fusion, as illustrated in Figure~\ref{fig:framework}. The key design principle is to integrate complementary inductive biases from convolutional and attention mechanisms before aligning features with the language model. This design explicitly decouples multimodal interaction from the LLM, allowing emotion-relevant representations to be formed in a controlled and task-aware manner.

Specifically, the Conv-Attention module consists of two parallel branches. The convolutional branch captures local and fine-grained patterns through restricted receptive fields, which are well suited for modeling short-term temporal dependencies and subtle modality-specific variations. In contrast, the attention branch emphasizes global interactions across modalities, enabling the model to identify emotionally salient components that may be spatially or temporally distant. By combining these two perspectives, the module provides a balanced representation that preserves both local detail and global context.
Before fusion, we employ a multilayer perceptron (MLP) to standardize the channel dimensionality of features extracted from different modalities, including audio, global visual, and temporal visual representations. Each modality-specific feature is represented in the form \pxj{$(\text{batch}, \text{depth}, \text{dim})$.} The normalized features are then organized into two hybrid feature structures, $\bm{F}_{d}$ and $\bm{F}_{s}$, defined as:
\begin{equation}
    \setlength\abovedisplayskip{3pt}
    \setlength\belowdisplayskip{3pt}
\label{eq:mlp_dimension}
\begin{aligned}
     &\bm{F}_{d} = \mathtt{Concat}(\mathtt{MLP}(u^{a}), \mathtt{MLP}(u^{v}_{glo}), \mathtt{MLP}(u^{v}_{temp})), \\ 
     &\bm{F}_{s} = \mathtt{Stack}(\mathtt{MLP}(u^{a}), \mathtt{MLP}(u^{v}_{glo}), \mathtt{MLP}(u^{v}_{temp}))
\end{aligned}
\end{equation}

In the attention branch, we apply $\mathtt{Attn\_MLP}(\cdot)$ to $\bm{F}_{d}$ to adapt its embedding depth to the sequence length of $\bm{F}_{s}$. A matrix product between the transformed $\bm{F}_{d}$ and $\bm{F}_{s}$ yields the attention-based fusion feature $\bm{F}_{attn}$:
\begin{equation}
    \setlength\abovedisplayskip{3pt}
    \setlength\belowdisplayskip{3pt}
\label{eq:attention}
\begin{aligned}
     \bm{F}_{attn} = \mathtt{Unsqueeze}(\mathtt{Attn\_MLP}(\bm{F}_{d}), \text{dim}=-1) \times \bm{F}_{s},
\end{aligned}
\end{equation}
where $\times$ denotes matrix multiplication and $\mathtt{Unsqueeze}(\cdot)$ expands the tensor dimension for alignment. This branch enables the model to dynamically emphasize emotionally informative modality components during fusion.

In parallel, the convolutional branch consists of an initial $\mathtt{Conv1d}(\cdot)$ layer followed by multiple residual convolution blocks equipped with the $\mathtt{Switch}(\cdot)$ activation function. Each block updates the feature representation as:
\begin{equation}
    \setlength\abovedisplayskip{3pt}
    \setlength\belowdisplayskip{3pt}
\label{eq:convolution}
\begin{aligned}
    \bm{F}_{conv}^{k} = \bm{F}_{conv}^{k-1} + \mathtt{Switch}(\mathtt{Conv1d}(\bm{F}_{conv}^{k-1})),
\end{aligned}
\end{equation}
where $k \,(k=1,2,\ldots,N)$ indexes the convolutional blocks. The residual structure promotes stable optimization, while the limited receptive fields introduce inductive biases that are particularly beneficial under limited-scale emotion datasets.
Finally, the outputs of the two branches, $\bm{F}_{conv}$ and $\bm{F}_{attn}$, are combined through element-wise addition to produce the final fused feature $u_{f}$. This fused representation serves as a compact yet expressive summary of multimodal emotional evidence, which is subsequently aligned with the language model through the modal adapter described in Section~\ref{sec:modal-adapter}. 

\subsection{Multimodal Alignment with Modal Adapter}
\label{sec:modal-adapter}
A central challenge in multimodal large language models lies in effectively aligning heterogeneous modality representations with the embedding space of a pre-trained language model. Audio, visual, and fused features are produced by encoders with fundamentally different architectures, temporal resolutions, and statistical properties. Directly injecting such heterogeneous features into an LLM without careful alignment often leads to representation mismatch, unstable optimization, or modality dominance, particularly in tasks that require fine-grained affective reasoning.

To address this issue, Emotion-LLaMAv2 introduces a \textit{modal adapter} that explicitly bridges the gap between upstream multimodal encoders and the language model. The modal adapter projects modality-specific feature sequences into a shared embedding space compatible with the LLM, enabling coherent cross-modal interaction and reasoning. Rather than relying on a single shared projection, we adopt modality-aware projections to preserve the semantic structure and temporal characteristics unique to each modality.

Formally, the modal adapter consists of a set of trainable linear projections $\boldsymbol{\sigma} = \{\mathcal{\sigma}^{a}, \mathcal{\sigma}^{v}_{glo}, \mathcal{\sigma}^{v}_{temp}, \mathcal{\sigma}^{f}\}$, each tailored to a specific upstream feature sequence. Given audio features $u^{a}$, global visual features $u^{v}_{glo}$, temporal visual features $u^{v}_{temp}$, and pre-fused multimodal features $u^{f}$, the adapter transforms each sequence into a corresponding token representation $\mathcal{T}$ in the LLM embedding space:
\begin{equation}
    \setlength\abovedisplayskip{3pt}
    \setlength\belowdisplayskip{3pt}
\mathcal{T} = \boldsymbol{\sigma} \cdot u, \quad
u \in \{u^{a}, u^{v}_{glo}, u^{v}_{temp}, u^{f}\}.
\label{eq:image_encoding}
\end{equation}

This design offers several advantages. First, modality-specific projections allow the adapter to account for differences in feature dimensionality, distribution, and semantic granularity across modalities, while still mapping all representations into a unified token space effectively. Second, operating at the token level preserves the temporal and structural information required for emotion understanding, in contrast to pooled or averaged representations that may obscure subtle affective cues crucially. Third, the lightweight linear design of the adapter avoids excessive parameter overhead, ensuring stable training when combined with LoRA-based fine-tuning of the language model consistently.

After projection, the resulting token sequences are concatenated according to the multimodal prompt template described in Section~\ref{sec:overview}. The final input stream to Emotion-LLaMAv2 therefore consists of ordered modality-aware tokens, including audio tokens $\langle\text{T}^{a}\rangle$, global visual tokens $\langle\text{T}^{v}_{glo}\rangle$, temporal visual tokens $\langle\text{T}^{v}_{temp}\rangle$, fusion tokens $\langle\text{T}^{f}\rangle$, and textual tokens $\langle\text{T}^{text}\rangle$. This explicit token composition enables the language model to attend selectively to different modalities through its cross-attention mechanism.

\subsection{The Perception-to-Cognition Training Framework}
\label{sec:training}
\pxj{In contrast to the Emotion-LLaMA framework, which utilized a coarse pre-training approach on MERR-Coarse followed by refinement on a limited set of human-verified MERR-Fine samples, Emotion-LLaMAv2 adopts a \textit{perception-to-cognition} training framework, which is a curriculum learning strategy that mirrors the developmental process of human emotional perception and cognitive reasoning.
The training begins with Basic Emotion Recognition, ensuring a smooth learning curve for the model by establishing foundational skills in identifying and categorizing emotions. This initial phase allows the model to grasp fundamental emotional signals effectively.
Following this, the second stage incorporates Emotion Reasoning, alongside multimodal emotion clues and contextual understanding. Here, the model enhances its ability to engage in higher-level cognitive processes, integrating visual, auditory, and linguistic evidence. This progression reflects the natural growth in human emotional development, where understanding emotions evolves from basic recognition to complex reasoning and contextual comprehension.
}

% \decrease{\sout{In contrast to the Emotion-LLaMA framework, which relied on coarse pre-training on MERR-Coarse followed by refinement on a limited set of human-verified MERR-Fine samples, Emotion-LLaMAv2 is trained on the substantially larger and higher-quality MMEVerse dataset with manually curated emotion annotations. The increased scale, modality diversity, and annotation fidelity of MMEVerse-Train enable a fundamentally different training paradigm, allowing the model to jointly learn emotion recognition and emotion reasoning within a single end-to-end multimodal large language model.
% Rather than treating recognition and reasoning as separate objectives or post-hoc capabilities, Emotion-LLaMAv2 adopts a unified multi-task instruction-tuning strategy. This design explicitly reflects the cognitive process of human affect understanding, where categorical perception and causal reasoning are tightly coupled. To this end, we employ a two-stage training scheme that progressively transitions the model from perception-level emotion alignment to higher-level multimodal emotional reasoning grounded in visual, auditory, and linguistic evidence.}}
% {\small
% }

\noindent \textbf{Stage 1--perception training phase: fundamental emotion alignment}.  
In the first stage, Emotion-LLaMAv2 is trained on MMEVerse-Train using emotion category labels only, without explicit reasoning supervision. Each training instance consists of multimodal feature tokens together with a randomly sampled instruction (\textbf{Box I} in supplement), accompanied by a recognition-specific task identifier. This identifier explicitly signals the model to perform categorical affect recognition, enabling consistent interpretation of the supervision signal across heterogeneous datasets listed in Table~\ref{tab:dataMMEVerse}.
The primary objective of this stage is to establish a robust alignment between multimodal representations and the discrete emotion label space. By repeatedly exposing the model to diverse visual, auditory, and linguistic manifestations of the same emotion category under varied instruction formulations, the model learns to associate affective concepts with modality-specific evidence in a stable and transferable manner. From a representation learning perspective, this stage anchors multimodal feature tokens into the word embedding space of the underlying language model, providing a well-conditioned initialization for subsequent reasoning-oriented training~\cite{chen2023minigpt, zhang2023video}. Empirically, we observe that this alignment stage significantly improves training stability and reduces spurious correlations during later reasoning supervision.

\noindent \textbf{Stage 2--cognition training phase: joint recognition and reasoning instruction tuning}.  
After completing Stage 1, Emotion-LLaMAv2 already demonstrates reliable emotion recognition performance across multiple datasets. To further enhance its ability to interpret emotional states and explain their underlying causes, we introduce a second instruction-tuning stage that jointly optimizes emotion recognition and multimodal reasoning. This stage leverages the rich descriptive annotations provided in MMEVerse-Train, which explicitly encode emotional cues, contextual factors, and cross-modal interactions.
During this phase, the model is trained with instructions randomly sampled (\textbf{Box II} in supplement). These prompts require the model not only to predict an emotion label but also to articulate a coherent reasoning process grounded in visual expressions, auditory prosody, and linguistic content. The supervision signal therefore operates at both the outcome level, namely the predicted emotion category, and the process level, namely the explanation that links multimodal evidence to affective interpretation. Through this joint optimization, Emotion-LLaMAv2 learns to integrate perception and reasoning within a unified autoregressive framework, enabling interpretable and context-aware emotion understanding.
Importantly, this two-stage curriculum training strategy avoids overfitting the model to surface-level descriptions or dataset-specific annotation artifacts. %Instead, it encourages the emergence of structured affective reasoning that generalizes across domains, modalities, and interaction scenarios. As demonstrated in Section~\ref{sec:experiments}, the resulting model exhibits strong robustness under modality degradation and superior generalization to unseen datasets, validating the effectiveness of the proposed training design.
\section{Experiments}
\label{sec:experiments}

\begin{table*}[ht]
\setlength{\abovecaptionskip}{1pt}
\setlength{\belowcaptionskip}{1pt}
\centering
\caption{\textbf{Comparison to state of the art on MMEVerse-Bench}. Note that Avg-9 \pxj{is identical to} the performance on MER-UniBench, while Avg-18 denotes the average performance across 18 test sets on MMEVerse-Bench. MELD-s refers to the sentiment recognition task, MAFW-m represents the multi-label emotion recognition task, and MC-EIU-i corresponds to the intent recognition task. ``Stage 2'' denotes the whole perception-to-cognition training framework. The best results are bold, and the second-best are highlighted in gray.}
\label{tab:emotionverse}
\resizebox{\textwidth}{!}{%
\renewcommand\arraystretch{1.2}
\begin{tabular}{l||cccccccccc}
\hline\thickhline
\rowcolor[RGB]{230, 230, 230}\textbf{Model} & \textbf{MER23}~\cite{lian2023mer} & \textbf{MER24}~\cite{lian2024mer} & \textbf{MELD} ~\cite{poria2018meld}& \textbf{IEMOCAP}~\cite{busso2008iemocap} & \textbf{MOSI}~\cite{zadeh2017tensor} & \textbf{MOSEI}~\cite{zadeh2018multimodal} & \textbf{SIMS}~\cite{yu2020ch} & \textbf{SIMS-v2}~\cite{liu2022make} & \textbf{OV-MERD+}~\cite{lian2025affectgpt} & \textbf{Avg-9} \\
\hline\hline
\multicolumn{11}{c}{\textit{\textbf{Input Modality: Visual, Text}}} \\
\hline
Video-LLaMA~\cite{zhang2023video}         &  38.65 & 44.40 & 31.60 & 20.36 & 69.19 & 36.75 & 31.16 & 31.21 &  49.91 & 39.25 \\
Video-LLaMA2~\cite{cheng2024videollama}        & 47.56 & 54.59 &  19.58 & 59.41 & 58.06 & 44.08 & 63.15 & 70.63 &  52.32 & 52.15 \\
Video-LLaVA~\cite{lin2023video}         & 54.82 & 43.31 & 35.96 & 37.17 & 71.31 & 40.81 & 61.20 & 57.17 &  52.50 & 50.47 \\
% GPT-4o(money)       & -- & -- & -- & -- & -- & -- & -- & --  & -- & -- \\
LLaMA3.2-90B-Vision~\cite{grattafiori2024llama3}& 58.49 & 62.72 & 48.82 & 57.71 & 82.06 & 65.74 & 77.65 & 80.26  & 54.09 & 65.28 \\
Qwen2.5-VL-72B~\cite{bai2025qwen2}     & 47.84 & 40.43 & 46.81 & 34.86 & 85.81 & 70.80 & 72.00 & 72.85  & 47.02 & 57.60 \\
\hline
\multicolumn{11}{c}{\textit{\textbf{Input Modality: Audio, Visual, Text}}} \\
\hline
Qwen2.5-Omni-7B~\cite{xu2025qwen2}        & 72.84 & 80.44 & 43.72 & 61.53 & 78.86 & 61.84 & 75.44 & 76.67  & 53.13 & 67.16 \\
HumanOmni-7B~\cite{zhao2025humanomni}        & 63.36 & 76.64 & 37.05 & 52.52 &  7.37 & 7.33 & 43.87 & 35.35  & 59.03 & 42.50 \\
HumanOmniV2-9B~\cite{yang2025humanomniv2}      & 68.59 & 57.41 & 42.74 & 60.95 & 76.56 & 75.44 & 83.58 & 81.65 & 61.62 & 67.62 \\
AffectGPT~\cite{lian2025affectgpt}           & \cellcolor[RGB]{231,231,231}\textbf{78.54} & \cellcolor[RGB]{245,245,245}78.80 & \cellcolor[RGB]{231,231,231}\textbf{55.65} & 60.54 & 81.30 & 80.90 & \cellcolor[RGB]{231,231,231}\textbf{88.49} & \cellcolor[RGB]{245, 245, 245}86.18  & 62.52 & 74.77 \\

Emotion-LLaMA~\cite{cheng2024emotion}      & 59.38 & 73.62 & 46.76 & 55.47 & 66.13 & 67.66 & 78.32 & 77.23  & 52.97 & 64.17 \\
% Emotion-LLaMAv2 (Stage1 w/o conv-attn) & 73.46& 84.59& 51.95& 83.82& 85.54& 84.92& 84.99& 85.77& 61.85& 77.43  \\
% Emotion-LLaMAv2 (joint)      & 71.27 & 77.94 & 42.20 & 82.25 & 88.80 & 85.45 & 84.81 & 83.22  & 63.88 & 75.54 \\
%  只保留一个token
% Emotion-LLaMAv2 (Stage1 with convattn) & 73.65& 83.47& 51.10& 81.08& 86.54& 86.31& 89.33& 86.13& 59.56& 77.46  \\
%  64个token
Emotion-LLaMAv2 (Stage 1) & \cellcolor[RGB]{245, 245, 245}77.28& \cellcolor[RGB]{231,231,231}\textbf{86.90}& \cellcolor[RGB]{245, 245, 245}51.32& \cellcolor[RGB]{231,231,231}\textbf{84.05}& \cellcolor[RGB]{245, 245, 245}86.28& \cellcolor[RGB]{231,231,231}\textbf{87.69}& \cellcolor[RGB]{245, 245, 245}87.72 & 86.02& \cellcolor[RGB]{245, 245, 245}62.90 & \cellcolor[RGB]{231,231,231}\textbf{78.91}  \\
%  64个token
% Emotion-LLaMAv2 (Stage2 w/o conv-attn) & 73.95& 84.86& 52.96& 81.26& 86.67& 86.07& 86.02& 86.64& 62.52& 77.88 \\
Emotion-LLaMAv2 (Stage 2) & 76.72& \cellcolor[RGB]{231,231,231}\textbf{86.90}& 48.61& \cellcolor[RGB]{245, 245, 245}83.60& \cellcolor[RGB]{231,231,231}\textbf{86.96}& \cellcolor[RGB]{245, 245, 245}86.15& 86.35 & \cellcolor[RGB]{231,231,231}\textbf{87.42}& \cellcolor[RGB]{231,231,231}\textbf{64.00} & \cellcolor[RGB]{245, 245, 245}78.52  \\
\hline\thickhline
% \bottomrule
\rowcolor[RGB]{230, 230, 230}\textbf{Model} & \textbf{CAER}~\cite{lee2019context} & \textbf{E3}~\cite{feng20243} & \textbf{DFEW}~\cite{jiang2020dfew} & \textbf{MAFW}~\cite{liu2022mafw} & \textbf{MAFW-m}~\cite{liu2022mafw} & \textbf{BOLD}~\cite{yang2023emoset} & \textbf{MELD-s}~\cite{poria2018meld} & \textbf{MC-EIU-i}~\cite{liu2024emotion} & \textbf{MC-EIU}~\cite{liu2024emotion} & \textbf{Avg-18}\\
\hline\hline
\multicolumn{11}{c}{\textit{\textbf{Input Modality: Visual, Text}}} \\
\hline
Video-LLaMA~\cite{zhang2023video}             & 37.62 & 32.47 & 37.59 & 24.36 & 22.53 &  7.58 & 51.84 & 22.62 & 44.83 & 35.26 \\
Video-LLaMA2~\cite{cheng2024videollama}            & 21.73 & 30.44 & 38.53 & 27.51 & 25.37 &  7.68 & 57.62 & 25.62 & 13.02 & 39.83 \\
Video-LLaVA~\cite{lin2023video}             & 27.87 & 31.24 & 37.29 & 22.57 & 22.03 &  7.07 & 43.14 & 22.02 & 30.61 & 38.78 \\
% GPT-4o(money)           & -- & -- & -- & -- & -- & -- & -- & --  & -- & -- \\
LLaMA3.2-90B-Vision~\cite{grattafiori2024llama3}           &37.92& 33.09& 46.90& 32.46& 21.56 & 7.17 & 63.83 & 37.86 & 43.49 & 50.66 \\
Qwen2.5-VL-72B~\cite{bai2025qwen2}     & 32.61 & 33.33 & 31.95 & 21.75 & 11.67 & 7.32 & 65.13 & 56.23  & 48.85 & 45.96 \\
\hline
\multicolumn{11}{c}{\textit{\textbf{Input Modality: Audio, Visual, Text}}} \\
\hline
Qwen2.5-Omni-7B~\cite{xu2025qwen2}        & 40.80 & 45.02 & 50.23 & 32.30 & 7.70 & 7.70 & 65.67 & 48.29  & 46.91 & 52.73\\
HumanOmni-7B~\cite{zhao2025humanomni}            & \cellcolor[RGB]{231,231,231}\textbf{62.12} & 41.33 & \cellcolor[RGB]{231,231,231}\textbf{74.33} & \textbf{75.42} & \cellcolor[RGB]{231,231,231}\textbf{44.07} & 8.45  & 38.77 & 30.42 & 52.54 & 45.00 \\
HumanOmniV2-9B~\cite{yang2025humanomniv2}          & 36.52 & 42.13 & 44.13 & 20.01 & 28.67 & 8.36  & 42.22 & 33.80 & 36.70 & 50.06 \\
AffectGPT~\cite{lian2025affectgpt}               & 30.45 & 27.61 & 43.61 & 26.26 & 23.18 & 7.22 & 58.93 & 37.95 & 46.35 & 54.11 \\
Emotion-LLaMA~\cite{cheng2024emotion}           & 40.54 & 35.42 & 46.86 & 30.67 & 26.11 & 7.67 & 59.54 & 38.09 & 48.11 & 50.59 \\
% Emotion-LLaMAv2 (Stage1 w/o conv-attn) & 57.34& 60.64& 70.78& 50.84& 36.17& 9.18& 74.48& 63.85& 62.19&65.13 \\
% Emotion-LLaMAv2 (joint)           & 56.62 & 60.27 & 69.59 & 48.56 & 33.79 & 8.55 & 70.46 & 63.67 & 64.22 & 64.20 \\ % 475.73
%  只保留1个token
% \rowcolor[RGB]{230, 230, 230}Emotion-LLaMAv2 (Stage1 with convattn) & 56.05& 59.16& 70.57 & 49.59& 35.23 & 9.23& 73.79 & 64.40& 61.82& 65.39\\ % 479.84
%  保留64个token
Emotion-LLaMAv2 (Stage 1) & 58.10& \cellcolor[RGB]{245, 245, 245}58.61& \cellcolor[RGB]{245, 245, 245}70.87 & 49.65& 34.06 & \cellcolor[RGB]{231,231,231}\textbf{9.21}&\cellcolor[RGB]{245, 245, 245}73.22 & \cellcolor[RGB]{245, 245, 245}63.43& \cellcolor[RGB]{245, 245, 245}61.50& \cellcolor[RGB]{245, 245, 245}66.05 \\ % 710.16
% Emotion-LLaMAv2 (Stage2 w/o conv-attn) & 57.83& 63.10& 70.61& 51.17& 41.35& 9.15& 75.02& 65.51& 61.82& 66.47\\ 
Emotion-LLaMAv2 (Stage 2) & \cellcolor[RGB]{245, 245, 245}58.21& \cellcolor[RGB]{231,231,231}\textbf{63.35}& 70.70 & \cellcolor[RGB]{245, 245, 245}50.73& \cellcolor[RGB]{245, 245, 245}35.88 & \cellcolor[RGB]{245, 245, 245}9.10& \cellcolor[RGB]{231,231,231}\textbf{74.71} & \cellcolor[RGB]{231,231,231}\textbf{65.60}& \cellcolor[RGB]{231,231,231}\textbf{64.36}& \cellcolor[RGB]{231,231,231}\textbf{66.63} \\ % 706.71
\hline\thickhline
\end{tabular}
    }
\end{table*}

\subsection{Implementation Details}
\label{sec:imple_details}
For the global visual feature, we extract the middle frame of the video and input it into the EVA-ViT-G model at a resolution of 448$\times$448 pixels. From the encoder output, we retain all patch tokens (excluding the [CLS] token) as fine-grained visual input for the LLM. For temporal visual features, 16 frames are uniformly sampled, processed by the same EVA model, and spatially pooled into 2$\times$2 embeddings per frame. For audio, the 16kHz mono stream is processed with Whisper-large-v3, and its variable-length output is normalized to a fixed 64-token sequence via 1D adaptive average pooling and zero-padding.

After feature extraction, cross-modal fusion is performed using the Conv-Attention module. This module jointly processes three streams: 1) the 64-token audio sequence; 2) the temporal visual sequence; and 3) a global image context. The global context is derived from the [CLS] token of the middle frame and broadcast to match 64-token sequence for alignment. The fusion output is a new sequence of multimodal features. All resulting embeddings are projected into a 4096-dimensional space via the modal adapter and concatenated with text tokens according to the multimodal prompt template, forming the input for Emotion-LLaMAv2.

% During the tuning process, we froze the visual and audio encoders, focusing on training the conv-attention module and modal adapter. For the language model (LLM), we utilize LLaMA2-chat (7B) equipped with LoRA for parameter-efficient tuning. Following the MiniGPT-v2~\cite{chen2023minigpt} approach, we fine-tune the query and value projection matrices ($\mathcal{W}_q$ and $\mathcal{W}_v$) by setting $r = 64$ and $\alpha = 16$. Consequently, the trainable parameters of Emotion-LLaMAv2 totaled only around 71 million, representing a mere 0.92\% of the overall parameter count. We train on 4*A100 GPUs for 300,000 steps, which takes around nine hours. Detailed information can be found on the project homepage and in the code repository.

Our model, Emotion-LLaMAv2, is built upon the MiniGPT-v2 framework~\cite{chen2023minigpt}. We freeze the pretrained visual and audio encoders, while the proposed conv-attention fusion module and modal adapters are trained from scratch. For the LLaMA2-chat (7B) LLM, we apply LoRA ($r=64$, $\alpha=16$) for parameter-efficient tuning of the query and value projection matrices ($\mathcal{W}_q$, $\mathcal{W}_v$). Training spans two stages over 100{,}000 steps, using the AdamW optimizer with 0.05 weight decay and a cosine learning rate schedule with 1000-step warmup to a peak of 1e-4. This setup yields 71 million trainable parameters (0.92\% of the total), with training completed in about three hours on 4 NVIDIA A100 GPUs. %Further details are provided on our project homepage \href{https://github.com/ooochen-30/Emotion-LLaMA-v2}{\textsf{https://github.com/ooochen-30/Emotion-LLaMA-v2}}.

\subsection{Evaluation Setup}
\label{sec:exp_setup}
% This section outlines the comprehensive experimental protocol used to evaluate Emotion-LLaMAv2. 
% For emotion recognition, we first detail our proposed MMEVerse-Bench and its multifaceted evaluation metrics. Subsequently, we describe the setup for assessing emotion reasoning, centered on the EMER ~\cite{lian2023explainable} dataset and its specific evaluation criteria. 
To verify the effectiveness of Emotion-LLaMAv2, we conducted extensive evaluations on MMEVerse-Bench for emotion recognition and EMER~\cite{lian2023explainable} for emotion reasoning. 

\noindent\textbf{Emotion Recognition Evaluation}. The evaluation protocol for MMEVerse-Bench ensures fair comparability by strictly following the metrics of MER-UniBench~\cite{lian2025affectgpt} for all overlapping datasets, while applying standard task-appropriate metrics for datasets unique to our collection. Specifically, we use hit rate for MER2023, MER2024, MELD-e, and IEMOCAP; weighted average F-score (WAF) for MOSI, MOSEI, SIMS, and SIMS-v2; and average F-score across emotion wheels for OV-MERD+. For our unique datasets, we adopt mean Average Precision (mAP) for multi-label tasks (MAFW-m and BOLD) and standard accuracy for the remaining datasets.

\noindent\textbf{Emotion Reasoning Evaluation}. 
The EMER dataset differs from traditional emotion datasets by including emotion trigger labels, such as facial micro-expressions, tone of speech, and video context, in addition to emotion categories. To evaluate the emotional reasoning capabilities of MLLMs on EMER, we employ GPT-4o to score predictions across three dimensions: (1) overlap between emotion-related clues, (2) overlap between summarized emotional states, and (3) completeness of the reasoning process across modalities. This multi-faceted evaluation provides a rigorous assessment of the models' ability to understand and explain emotions in a multimodal context. Prompts for emotional reasoning and overlap calculation are listed in the supplement \textbf{Box III}.

\subsection{Main Results}
In this section, we present the main quantitative results of Emotion-LLaMAv2 on the emotion recognition and reasoning benchmarks.

\noindent\textbf{Emotion Recognition with Emotion-LLaMAv2}. 
We evaluate Emotion-LLaMAv2's emotion recognition performance against mainstream vision-language models and recent Omni models. For fairness, all compared models are run with their default weights and preprocessors. As shown in Table~\ref{tab:emotionverse}, our model achieves state-of-the-art results on both MER-UniBench and MMEVerse-Bench. It obtains 66.63\% on MMEVerse-Bench, outperforming the next-best model (AffectGPT) by about 12\%. On MER-UniBench, Emotion-LLaMAv2 reaches 78.91\% accuracy, exceeding AffectGPT by roughly 4\%. Notably, AffectGPT's zero-shot capability on MER-UniBench did not generalize well to new datasets in MMEVerse-Bench, underscoring the broader evaluation scope of our benchmark. 

HumanOmni-7B performs particularly well on datasets such as DFEW, MAFW, and CAER, likely due to its training focus on these datasets with GPT4o-reformulated question-answer annotations. Most models show limited performance on the multi-label tasks MAFW-m and BOLD, reflecting the difficulty and limited adoption of these emotion labels. Finally, while Stage 2 training slightly improves Emotion-LLaMAv2, reasoning-focused tuning has limited impact on recognition since it emphasizes matching complex descriptions rather than categorical accuracy.

\begin{table}[t]
\centering
\caption{Comparison of \textbf{multimodal emotion reasoning results on the EMER dataset}, with Clue Overlap and Label Overlap scores ranging from 0 to 10. The scores differ from our preliminary version~\cite{cheng2024emotion} as GPT-4-o replaces GPT-3.5-turbo. These results are valid only when evaluated using a consistent GPT version to ensure fairness and reliability.}
% \resizebox{\linewidth}{!}{
\renewcommand\arraystretch{1.15}
% \tiny
\begin{tabular}{l||cc}
\thickhline
\rowcolor[RGB]{230, 230, 230}\textbf{Models} & \textbf{Clue Overlap} & \textbf{Label Overlap} \\
\hline\hline
\rowcolor[RGB]{245, 245, 245}PandaGPT~\cite{su2023pandagpt}            & 4.86 & 4.39 \\
Valley~\cite{luo2023valley}              & 4.32 & 5.45 \\
\rowcolor[RGB]{245, 245, 245}VideoChat-Embed~\cite{li2023videochat}     & 4.04 & 5.12 \\
VideoChat-Text~\cite{li2023videochat}      & 3.28 & 3.89 \\
\rowcolor[RGB]{245, 245, 245}Video-ChatGPT~\cite{maaz2023video}       & 4.81 & 5.07 \\
Video-LLaMA~\cite{zhang2023video}         & 3.68 & 5.01 \\
\rowcolor[RGB]{245, 245, 245}Qwen2.5-Omni~\cite{xu2025qwen2}        & 5.60 & 5.04 \\
Video-LLaMA2~\cite{cheng2024videollama}        & 5.63 & 5.51 \\
\rowcolor[RGB]{245, 245, 245}AffectGPT~\cite{lian2025affectgpt}           & 5.87 & 5.79 \\
Emotion-LLaMA~\cite{cheng2024emotion}       & 5.89 & 6.89 \\
\hline
\rowcolor[RGB]{245, 245, 245}Emotion-LLaMAv2 & \textbf{7.30} & \textbf{7.14} \\ % (Stage 2)
\thickhline
\end{tabular}
\label{tab:emer}
\end{table}

\begin{table}[t]
\centering
\caption{\textbf{Overall performance comparison ($\%$) on recent MME-Emotion~\cite{zhang2025mme}.} 
 For both open-source and closed-source models, the best results are bold, and the second-best are highlighted in gray.}
\label{tab:mme}
\resizebox{\columnwidth}{!}{
\begin{tabular}{l||c||ccccc}
\thickhline
\rowcolor[RGB]{230, 230, 230}\textbf{Model} & \textbf{LLM Size} & \textbf{Avg Step} & \textbf{Avg Token} & \textbf{Rec-S} & \textbf{Rea-S} & \textbf{CoT-S} \\
\midrule
\multicolumn{7}{c}{Open-source MLLMs} \\
\midrule
Qwen2-Audio \cite{chu2024qwen2} & 7B & 3.0 & 40.3 & 34.1 & 50.4 & 42.3 \\
Audio-Reasoner \cite{xie2025audio} & 7B & 5.0 & 356.8 & \cellcolor[RGB]{245, 245, 245}38.1 & 71.6 & \cellcolor[RGB]{245, 245, 245}54.8 \\
% Qwen2-VL-7B \cite{wang2024qwen2} & 7B &   2.9 & 68.0 & 29.2 & 38.1 & 33.7 \\
% Qwen2-VL-72B \cite{wang2024qwen2} & 72B &  1.5 & 24.8 & 31.1 & 10.5 & 20.8 \\
Qwen2.5-VL-7B \cite{bai2025qwen2} & 7B &  4.8 & 169.7 & 28.4 & 64.8 & 46.6 \\
Qwen2.5-VL-72B \cite{bai2025qwen2} & 72B &  4.8 & 266.3 & 31.3 & \cellcolor[RGB]{231,231,231}\textbf{75.7} & 53.5 \\
QVQ \cite{team5qvq} & 72B & 5.5 & 899.9 & 31.4 & 70.1 & 50.8 \\
Video-LLaVA \cite{lin2023video} & 7B & 2.3 & 19.1 & 25.8 & 32.8 & 29.3 \\
Video-LLaMA \cite{zhang2023video} & 7B & 4.5 & 122.5 & 26.1 & 48.5 & 37.3 \\
Video-LLaMA2 \cite{cheng2024videollama} & 7B & 2.6 & 37.7 & 29.2 & 27.7 & 28.4 \\
Qwen2.5-Omni \cite{xu2025qwen2} & 7B & 3.7 & 78.6 & 17.4 & 59.3 & 38.4 \\
Emotion-LLaMA \cite{cheng2024emotion} & 7B &1.0 & 2.3 & 25.1 & 0.4 & 12.8 \\
HumanOmni \cite{zhao2025humanomni} & 7B & 1.0 & 1.3 & 36.0 & 0.3 & 18.1 \\
R1-Omni \cite{zhao2025r1omni} & 0.5B & 5.0 & 156.2 & 26.3 & 58.6 & 42.4 \\
AffectGPT \cite{lian2025affectgpt} & 7B &  4.9 & 122.8 & 11.9 & 50.6 & 31.2 \\
Emotion-LLaMAv2 (Stage 2) & 7B &  5.8 &184.5 & \cellcolor[RGB]{231,231,231}\textbf{43.26} & \cellcolor[RGB]{245, 245, 245}71.63 & \cellcolor[RGB]{231,231,231}\textbf{57.45} \\
\midrule
\multicolumn{7}{c}{Closed-source MLLMs} \\
\midrule
GPT-4o \cite{hurst2024gpt} & \textemdash &  4.4 & 169.4 & 27.8 & \cellcolor[RGB]{231,231,231}\textbf{79.8} & \cellcolor[RGB]{245, 245, 245}53.8 \\
GPT-4.1 \cite{openaigpt41} & \textemdash & 5.2 & 141.2 & 28.8 & 65.2 & 47.0 \\
Gemini-2.0-Flash \cite{gemini20flash} &\textemdash & 4.1 & 64.7 & \cellcolor[RGB]{245, 245, 245}36.3 & 60.0 & 48.1 \\
Gemini-2.5-Flash \cite{gemini25flash} & \textemdash& 4.3 & 261.8 & 34.7 & 52.7 & 43.7 \\
Gemini-2.5-Pro \cite{gemini25pro} & \textemdash &  5.1 & 538.6 & \cellcolor[RGB]{231,231,231}\textbf{39.3} & \cellcolor[RGB]{245, 245, 245}72.7 & \cellcolor[RGB]{231,231,231}\textbf{56.0} \\
\bottomrule
\end{tabular}%
}
\end{table}

\begin{table}[t]
\setlength{\abovecaptionskip}{1pt}
\setlength{\belowcaptionskip}{1pt}
\centering
\caption{The Emotion-LLaMAv2 performance trained on different MER instruction tuning datasets.}
\resizebox{\linewidth}{!}{
\renewcommand\arraystretch{1.15}
\scriptsize
\begin{tabular}{l||cc}
\thickhline
\rowcolor[RGB]{230, 230, 230}\textbf{Training Dataset} & \textbf{MER-Unibench}~\cite{lian2025affectgpt} & \textbf{MMEVerse-Bench} \\
\hline\hline
MERR-Coarse~\cite{cheng2024emotion}            & 61.78 & 44.42 \\
MERR-Fine~\cite{cheng2024emotion}              & 63.08 & 48.72 \\
MER-Caption+~\cite{lian2025affectgpt}           & 69.02 & 50.44 \\
\hline
\rowcolor[RGB]{245, 245, 245}MMEVerse               & \textbf{78.91}& \textbf{66.05} \\
\thickhline
\end{tabular}}
\label{tab:ablation_datasets}
\end{table}

\begin{figure}[t]
\setlength{\abovecaptionskip}{1pt}
\setlength{\belowcaptionskip}{1pt}
    \centering
    \includegraphics[width=\linewidth]{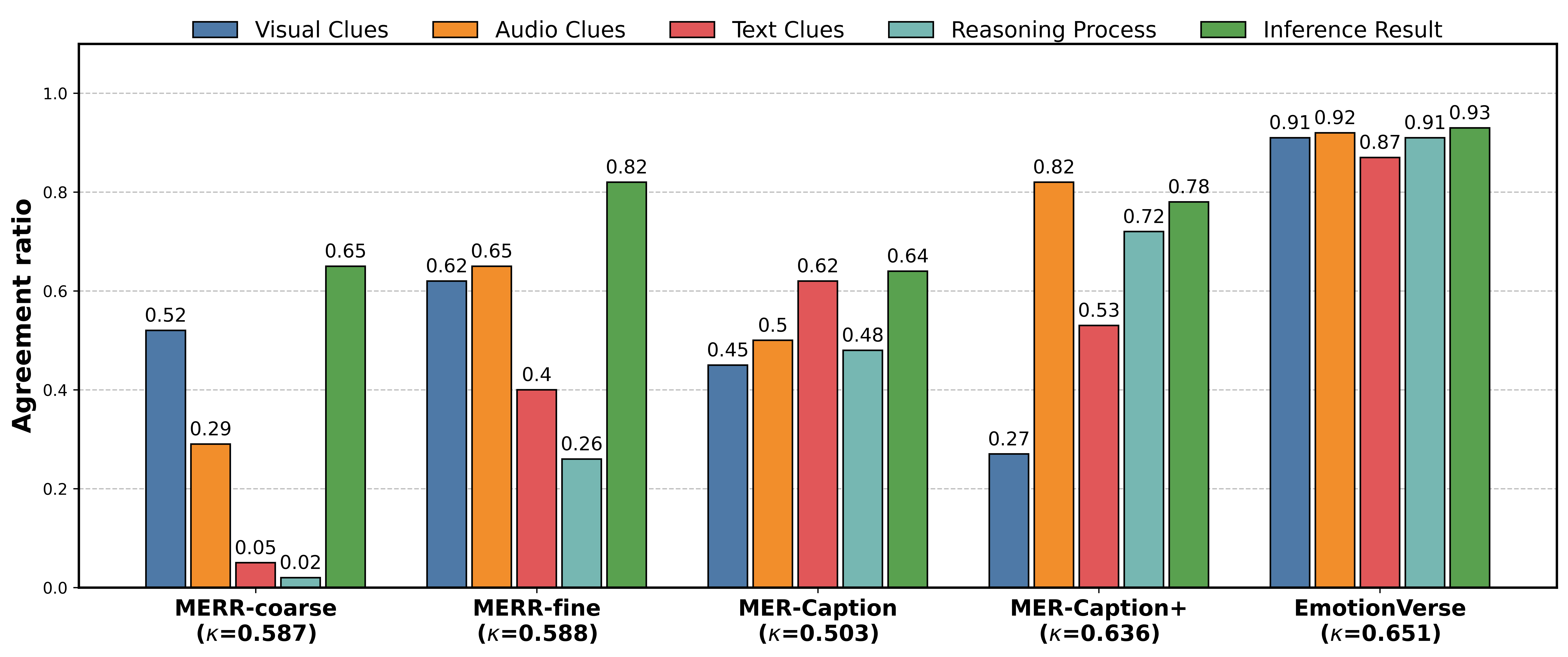}
    \caption{User Study on Data Quality.}
    \label{fig:humancheck}
    % \vspace{-4mm}
\end{figure}

\begin{table}[t]
\setlength{\abovecaptionskip}{1pt}
\setlength{\belowcaptionskip}{1pt}
\centering
\caption{Ablation study of the Conv-Attention module.}
\label{tab:Conv-AttnEval}
\setlength{\tabcolsep}{6pt}
\resizebox{\linewidth}{!}{
\renewcommand\arraystretch{1.15}
\footnotesize
\begin{tabular}{l||cc cc}
\thickhline
\rowcolor[RGB]{230, 230, 230}& \multicolumn{2}{c}{\textbf{MER-UniBench}~\cite{lian2025affectgpt}} & \multicolumn{2}{c}{\textbf{MMEVerse-Bench}} \\
\cline{2-5}
\rowcolor[RGB]{230, 230, 230}\textbf{Conv-Attn} & \multicolumn{1}{c}{\textcolor{red}{\xmark}} & \multicolumn{1}{c}{\mygreencheck} & \multicolumn{1}{c}{\textcolor{red}{\xmark}} & \multicolumn{1}{c}{\mygreencheck} \\
\hline\hline
\textbf{Stage 1} & 77.43 & 78.91 \increase{1.48} & 65.13 & 66.05 \increase{0.92} \\
\textbf{Stage 2} & 77.88 & 78.52 \increase{0.64} & 66.47 & 66.63 \increase{0.16} \\
\thickhline
\end{tabular}}
\end{table}

\begin{table}[t]
\setlength{\abovecaptionskip}{1pt}
\setlength{\belowcaptionskip}{1pt}
\centering
\caption{Comparison of different pre-fusion architectures.}
\setlength{\tabcolsep}{6pt}
\resizebox{\linewidth}{!}{
\renewcommand\arraystretch{1.15}
\footnotesize
\begin{tabular}{c||cc}
\thickhline
\rowcolor[RGB]{230, 230, 230}\textbf{Pre-fusion} & \textbf{MER-Unibench}~\cite{lian2025affectgpt} & \textbf{MMEVerse-Bench} \\
\hline\hline
\textcolor{red}{\xmark}               & 77.43    & 65.13 \\
AffectGPT-Q-Former~\cite{lian2025affectgpt}      & 77.95\increase{0.52}    & 65.85\increase{0.72} \\
AffectGPT-Attn~\cite{lian2025affectgpt}    & 77.65\increase{0.22}    & 65.37\increase{0.24} \\
Our Attn     & 78.05\increase{0.62}    & 65.68\increase{0.55} \\
Our Conv     & 77.93\increase{0.50}    & 65.92\increase{0.79} \\
\hline
\rowcolor[RGB]{245, 245, 245}Conv-Attn               & \textbf{78.91} \increase{1.48} & \textbf{66.05} \increase{0.92} \\
\thickhline
\end{tabular}}
\label{tab:prefusion}
\vspace{-0.2cm}
\end{table}

\begin{table}[t]
\setlength{\abovecaptionskip}{1pt}
\setlength{\belowcaptionskip}{1pt}
\centering
\caption{Comparison of different training schemes.}
\label{tab:train}
\setlength{\tabcolsep}{6pt}
\resizebox{\linewidth}{!}{
\renewcommand\arraystretch{1.15}
\footnotesize
\begin{tabular}{c||cc}
\thickhline
\rowcolor[RGB]{230, 230, 230}\textbf{Training} & \textbf{MER-Unibench}~\cite{lian2025affectgpt} & \textbf{MMEVerse-Bench} \\
\hline\hline
Perception (Stage 1)            & 78.91    & 66.05 \\
Single joint training   & 75.54   & 64.20 \\
\hline
\rowcolor[RGB]{245, 245, 245}P.-to-C. (Stage 2)     & 78.52   & 66.63 \\
\thickhline
\end{tabular}}
\end{table}

\noindent\textbf{Emotion Reasoning with Emotion-LLaMAv2}. 
We present results for the emotion reasoning task in Table~\ref{tab:emer}. Analysis of baseline models yields several insights. For example, VideoChat shows that aligning visual features directly with the language embedding space ("Embed" variant) is far more effective than converting them into text tokens ("Text" variant). This observation validates our architectural choice of projecting multimodal features into the language space.

AffectGPT, our primary competitor, also demonstrates strong reasoning capabilities, highlighting the effectiveness of its pre-fusion module and mixed-task instruction tuning. Emotion-LLaMAv2 (Stage 2) establishes a new state-of-the-art, achieving the highest scores across all key criteria (e.g., 7.30 on Clue Overlap and 7.14 on Label Overlap), clearly surpassing other models. These results confirm the effectiveness of our two-stage training strategy: first building a robust recognition foundation, then dedicating the model to complex end-to-end reasoning.

\noindent\textbf{Results on MME-Emotion}. MME-Emotion~\cite{zhang2025mme} is a very recent holistic evaluation benchmark designed for emotion recognition and reasoning. In Table \ref{tab:mme}, we present a comparison of our model, Emotion-LLaMAv2, with both open-source and closed-source multimodal LLMs using the standard evaluation metric from the benchmark. Remarkably, our model outperforms all others in both the recognition score (Rec-S) and Chain-of-Thought score (CoT-S). Additionally, it ranks second in the reasoning score (Rea-S) among open-source models. Notably, all closed-source multimodal LLMs rely solely on visual and textual information yet maintain competitive performance. Our state-of-the-art results on MME-Emotion indicate that advanced training on a large, multimodal emotion understanding dataset can surpass even commercialized large models trained on extensive datasets. 
% We attribute this superior performance directly to our structured, two-stage training curriculum. By first building a robust recognition foundation before dedicating the model to complex, end-to-end reasoning, our approach cultivates a deeper causal understanding than a single-stage, mixed-task paradigm. This is particularly reflected in the higher Clue Overlap score, indicating more coherent and causally-grounded explanations.

% \begin{figure}
%     \centering
%     \includegraphics[width=0.5\textwidth]{images/dataEval.png}
%     \caption{\textcolor{red}{Effectiveness of MMEVerse}}
%     \label{fig:dataEval}
% \end{figure}

\subsection{Data Evaluation}
\noindent\textbf{Effectiveness of MMEVerse}. Table \ref{tab:ablation_datasets} presents an evaluation of recent multimodal emotion instruction tuning datasets, including MERR-Coarse, MERR-Fine, and MER-Caption+. We train Emotion-LLaMAv2 on these datasets with their emotion categories and showcase its performance on both MER-UniBench and MMEVerse-Bench. The results reveal that training on MMEVerse-Train yields significantly better performance, particularly on MMEVerse-Bench. This can be attributed to two main factors: its data diversity and annotation reliability.
On the one hand, its diverse sources, spanning from YouTube videos to interviews, capture more authentic "in-the-wild" emotional expressions, equipping the model with superior generalization capabilities. On the other hand, MMEVerse is more accurate in annotation quality since each emotion label is manually annotated, enabling the model to learn true multimodal-emotion correlations without label noise.

% \pxj{As shown in Fig~\ref{fig:humancheck}, we conducted a human study to validate data quality across five dimensions. EmotionVerse significantly outperforms all benchmarks, achieving over 0.91 average "Yes" proportion in most categories, particularly in Reasoning Process and Inference Result. To ensure reliability, we measured the inter-rater agreement using Cohen’s Kappa ($\kappa$). EmotionVerse achieves a $\kappa$ of 0.651, indicating substantial agreement between annotators. These results demonstrate that EmotionVerse provides not only higher-quality multi-modal clues but also more reliable logical reasoning compared to existing datasets.}

\noindent\textbf{User Study on Data Quality}. A user study was conducted by selecting 50 samples from each dataset for human verification, yielding critical insights into their quality. As shown in Figure~\ref{fig:humancheck}, MMEVerse demonstrates the highest reliability across all modalities, with agreement rates exceeding 0.90 for visual, audio, text clues, reasoning processes, and inference results, showcasing its strong alignment with human perception. Among the existing MER datasets, MER-Caption+ outperforms others due to its multi-level model-based filtering, enhancing interpretability and reasoning. In contrast, MER-coarse and MER-fine exhibit moderate agreement, particularly in audio clues and reasoning processes, highlighting their limitations in delivering sufficient multimodal context. Additionally, some unimodal clues score relatively lower due to unrelated emotional information they carry, further emphasizing the need for improved dataset design. The study also demonstrates the effectiveness of MMEVerse for instruction tuning, solidifying its value in advancing multimodal research.

\pxj{To further assess the quality of these datasets, we utilize Cohen’s Kappa ($\kappa$) coefficient to analyze the agreement between two experts based on the five perspectives in Figure~\ref{fig:humancheck}. Following the initial human verification, we asked an additional expert to conduct a review, allowing us to compute the Cohen’s Kappa ($\kappa$) coefficient. As illustrated at the bottom of Figure~\ref{fig:humancheck}, EmotionVerse achieves a $\kappa$ of 0.651, indicating substantial agreement between annotators.}

\subsection{Ablation Studies}

\noindent To validate the key design choices of Emotion-LLaMAv2, we conduct ablation studies on its core components and training data. For this evaluation, only Stage~1 training is executed, focusing on emotion recognition.

\noindent\textbf{Effectiveness of the Conv-Attention Module}. 
Table~\ref{tab:Conv-AttnEval} reports the evaluation of the Conv-Attention module. It significantly improves Stage~1 performance, increasing the average score by 1.48\% on MER-UniBench and 0.92\% on MMEVerse-Bench. This highlights its importance: pre-fusion representations complement unimodal embeddings through interaction before feeding into the LLM. For models trained in Stage~2, the gains remain but are slightly reduced, as complex reasoning training can offset some improvements.

To further assess the pre-fusion module, we compare our Conv-Attn architecture against the Q-Former and the MLP-based attention mechanism from AffectGPT~\cite{lian2025affectgpt}. As shown in Table~\ref{tab:prefusion}, the inclusion of any pre-fusion module consistently improves performance over the baseline, validating the strategy of refining modal features prior to joint processing by the LLM. Among individual architectures, both the Q-Former and our Conv model perform strongly, suggesting that preserving local temporal information during pre-fusion is particularly effective. Ultimately, Conv-Attn achieves the highest performance, reflecting a synergistic effect between its components: convolutional layers excel at extracting local temporal features, while attention mechanisms capture long-range dependencies and global context. By integrating these complementary strengths, the model produces richer and more comprehensive representations, leading to superior final performance.

\pxj{
\noindent\textbf{Effectiveness of the perception-to-cognition training scheme}. 
Table~\ref{tab:train} presents an evaluation of different strategies for instruction tuning. While one might argue that a joint training scheme for recognition and reasoning is simpler and more efficient, akin to our preliminary Emotion-LLaMA, surprisingly, this straightforward approach yields significantly poorer performance on both benchmarks listed in Table~\ref{tab:train}. This can be attributed to the difficulty of learning both perception and cognition abilities from scratch, in contrast to a curriculum learning approach.
}

% This richer, fused feature stream acts as a powerful supplement to the original single-modality features, providing the model with a more comprehensive and discriminative basis for accurate emotion recognition. The benefits of this enhanced, multi-faceted input persist into Stage 2, particularly on the more complex MMEVerse-Bench (+0.58\%). Overall, the results validate that our strategy of complementing unimodal pathways with a dedicated fusion module is highly effective.

\begin{table}[t]
\centering
\caption{Ablation study on \textbf{Modality Encoders}. \pxj{All the models are trained on MMEVerse-Train, and their performance is reported on MMEVerse-Bench.}} % Ablation study on unimodal encoders and multimodal fusion. Performance is measured solely on the MMEVerse-Bench dataset.
\resizebox{\linewidth}{!}{
\renewcommand\arraystretch{1.2}
\scriptsize
    \begin{tabular}{ccc||c}
       \thickhline  
\rowcolor[RGB]{230, 230, 230}    \textbf{Audio Encoder} & \textbf{Image Encoder} & \textbf{Video Encoder} & \textbf{Avg}\\
    \hline\hline
\multicolumn{4}{c}{\textit{\textbf{Unimodal Encoder}}} \\
    \hline
\rowcolor[RGB]{245, 245, 245}    Whisper~\cite{radford2023robust} & -                   & -           & {56.28}\\
    HuBERT~\cite{hsu2021hubert}  & -                   & -           & \textbf{58.92}\\
\rowcolor[RGB]{245, 245, 245}    -       & EVA~\cite{fang2023eva}                 & -           & {57.70}\\
    -       & MAE~\cite{he2022masked}                 & -           & {50.92}\\
\rowcolor[RGB]{245, 245, 245}    -       & CLIP~\cite{radford2021learning}                & -           & {57.53}\\
    -       & -                   & VideoMAE~\cite{tong2022videomae}    & {53.39}\\
\rowcolor[RGB]{245, 245, 245}    -       & -                   & VideoMAEv2~\cite{wang2023videomae}  & {50.09}\\
    -       & -                   & EVA~\cite{fang2023eva}       & {56.52}\\
\rowcolor[RGB]{245, 245, 245}    -       & -                   & CLIP~\cite{radford2021learning}        & {55.69}\\
    \hline
\multicolumn{4}{c}{\textit{\textbf{Two Encoders}}} \\
    \hline 
\rowcolor[RGB]{245, 245, 245}    HuBERT~\cite{hsu2021hubert}  & EVA~\cite{fang2023eva}                 & -               & {60.62}\\
    Whisper~\cite{radford2023robust} & EVA~\cite{fang2023eva}                 & -               & \textbf{61.73}\\
\rowcolor[RGB]{245, 245, 245}    HuBERT~\cite{hsu2021hubert}  & -                   & CLIP~\cite{radford2021learning}            & {57.88}\\
    HuBERT~\cite{hsu2021hubert}  & -                   & EVA~\cite{fang2023eva}           & {59.31}\\
\rowcolor[RGB]{245, 245, 245}    Whisper~\cite{radford2023robust} & -                   & CLIP~\cite{radford2021learning}            & {55.76}\\
    Whisper~\cite{radford2023robust} & -                   & EVA~\cite{fang2023eva}           & {60.67}\\
    % -       & EVA                 & CLIP        & {}\\
    % -       & EVA                 & EVA64       & {}\\
    \hline
\multicolumn{4}{c}{\textit{\textbf{Three Encoders}}} \\
    \hline
\rowcolor[RGB]{245, 245, 245}    HuBERT~\cite{hsu2021hubert}  & EVA~\cite{fang2023eva}                 & CLIP~\cite{radford2021learning}            & {63.66}\\
    HuBERT~\cite{hsu2021hubert}  & EVA~\cite{fang2023eva}                 & EVA~\cite{fang2023eva}           & {61.90}\\
\rowcolor[RGB]{245, 245, 245}    Whisper~\cite{radford2023robust} & EVA~\cite{fang2023eva}                 & CLIP~\cite{radford2021learning}            & {62.46}\\
    Whisper~\cite{radford2023robust} & EVA~\cite{fang2023eva}                 & EVA~\cite{fang2023eva}           & \textbf{66.05}\\  
    \hline
\multicolumn{4}{c}{\textit{\textbf{Emotion-LLaMA Setup}}} \\
    \midrule
    % HuBERT  & EVA                 & MAE+VideoMAE   & {60.55}\\
\rowcolor[RGB]{245, 245, 245}    HuBERT~\cite{hsu2021hubert}  & EVA~\cite{fang2023eva}                 & MAE~\cite{he2022masked}+VideoMAE~\cite{tong2022videomae}    & {61.44}\\
    \thickhline
    \end{tabular}}
\label{tab:encFusionEval}
  \end{table}

\begin{figure*}[t]
\setlength{\abovecaptionskip}{1pt}
\setlength{\belowcaptionskip}{1pt}
    \centering
    \subfloat[]{
    \includegraphics[width=0.3\linewidth]{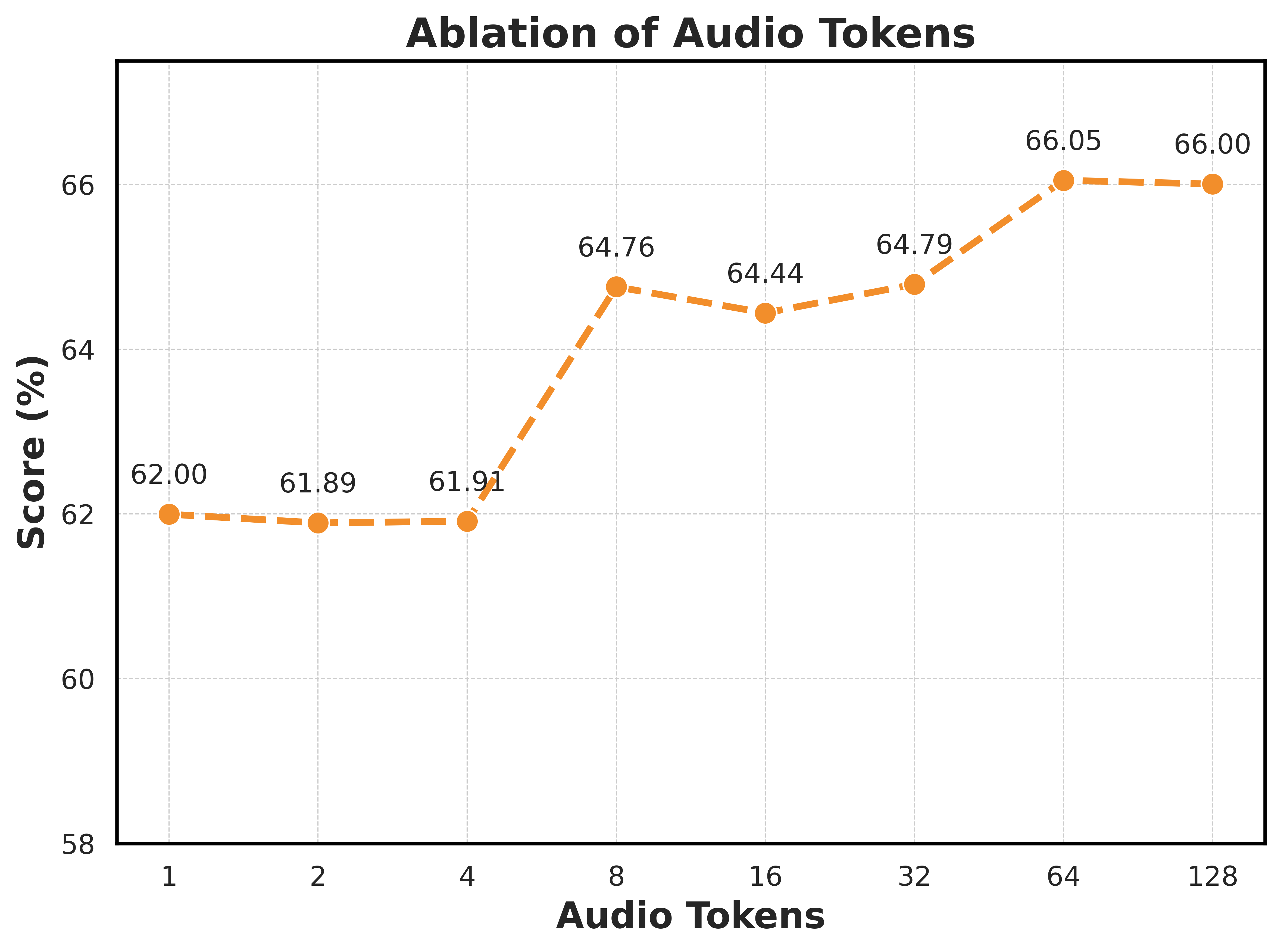}  
    }
    \subfloat[]{
    \includegraphics[width=0.3\linewidth]{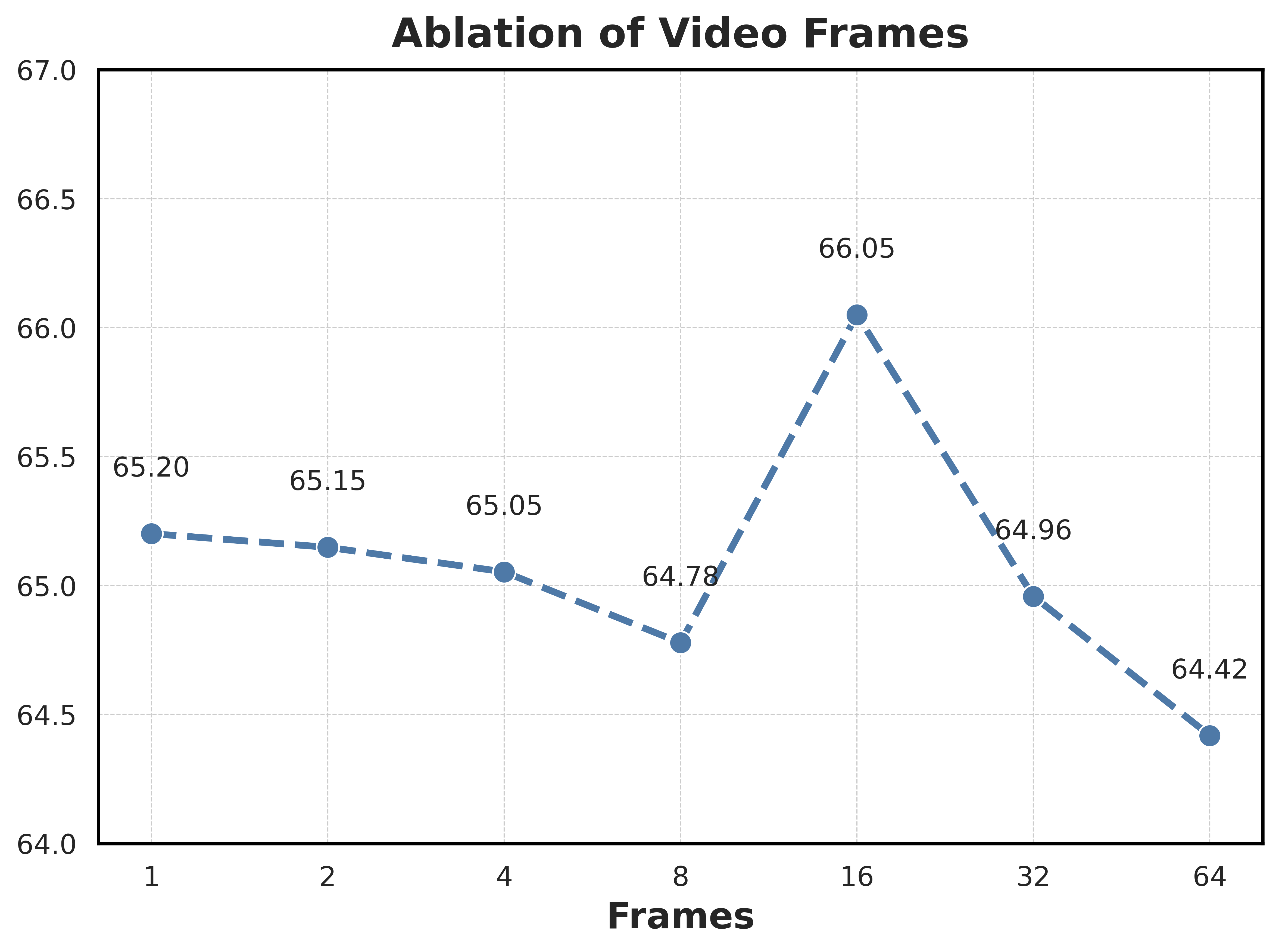}  
    }
    \subfloat[]{
    \includegraphics[width=0.3\linewidth]{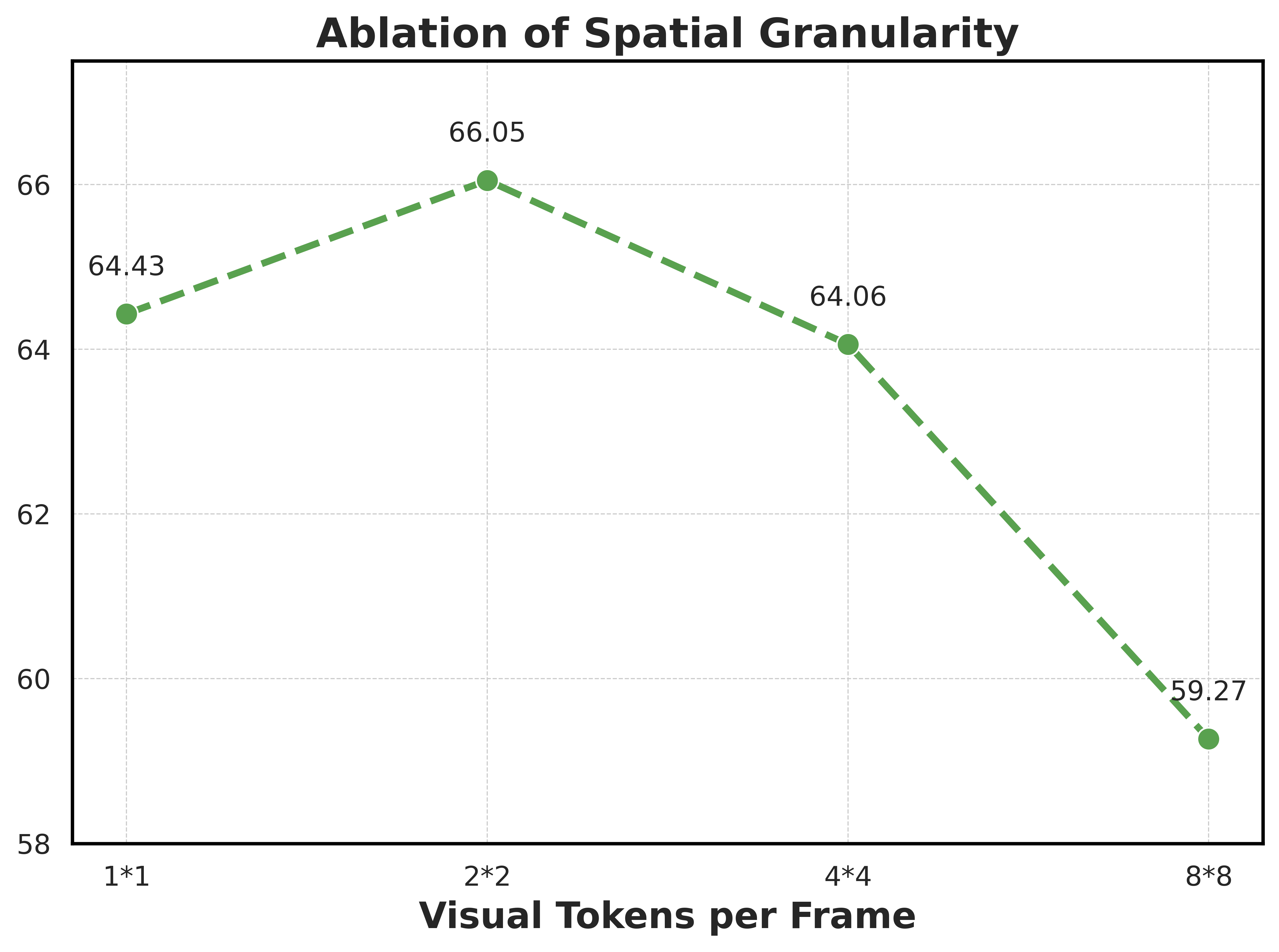}  
    }
    \caption{Ablation of \textbf{temporal representations}, mainly evaluated on MMEVerse-Bench. We analyze the model's performance as a function of: (a) number of audio tokens, (b) temporal sampling density (number of video frames), and (c) spatial granularity of each frame (number of visual tokens). We default to sampling 16 frames per video and 64 frames per audio.}
    \label{fig:TemporalRepresentations}
\vspace{-0.3cm}
\end{figure*}
\begin{table*}[ht]
\setlength{\abovecaptionskip}{1pt}
\setlength{\belowcaptionskip}{1pt}
\caption{An example of multimodal emotion reasoning comparing our Emotion-LLaMAv2 with other MLLMs. Incorrect reasoning is marked in red, correct reasoning in blue, and hallucinations in green. }
\centering  
% \vspace{-2mm}
% \scalebox{0.70}{
\resizebox{\linewidth}{!}{
\renewcommand\arraystretch{1.15}
\scriptsize
\begin{tabular}{l||p{14.5cm}}
\hline\thickhline
\rowcolor[RGB]{230, 230, 230}\multicolumn{2}{c}{\bf~~~~~~~~~~~An Example of Multimodal Emotion Reasoning} \\
\hline\hline
& {\includegraphics[height=2.01cm]{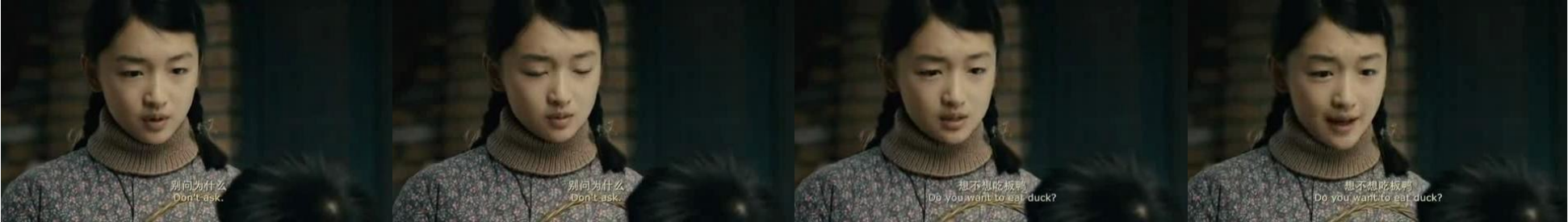}} \\
Subtitle (En) & {Don't ask why, do you want to eat Peking duck?} \\
\hline
Ground Truth & {In the video, she seemed a bit unhappy with the other person's questions, possibly because they asked some sensitive questions. Her speech was fast and she slightly frowned. Therefore, it is inferred that her emotional state is anger.}  \\
\hline
\makebox[0pt][l]{Valley~\cite{luo2023valley}} & {From the video, we can infer that the little girl is likely in a playful or mischievous mood. In one scene, she is seen making a funny face, which suggests that \textcolor{red}{she is having fun and feels at ease}. In addition, \textcolor{green}{when she is shown eating a donut}, she seems to be enjoying her treat. Overall, the little girl appears to be in a happy and energetic emotional state.
}  \\
% % \midrule
% % VideoChat-Embed & {}  \\
\hline
\makebox[0pt][l]{PandaGPT~\cite{su2023pandagpt}} & {The person in the video appears to be smiling, which suggests they are in a positive and happy emotional state. The smiling person is holding a sign that reads, ``Don’t ask why, do you want to eat Peking duck?'' This statement implies that the person is \textcolor{red}{enthusiastic and excited} about Peking duck, which may represent a delicious dish or a special occasion. Their cheerful expression and the message on the sign indicate that they are likely enjoying themselves and sharing their enthusiasm with others.}  \\
\hline
\makebox[0pt][l]{Video-LLaMA2~\cite{cheng2024videollama}} & {The person in the video appears to be in a calm and composed state. They are asking a question in a soft and gentle tone, without any signs of aggression or frustration. The use of the word ``Peking duck'' suggests that \textcolor{red}{they are being playful and humorous}, which could indicate a positive emotional state. Overall, it seems that the person is simply seeking an answer to their question, without any underlying emotional issues.}  \\
\hline
\makebox[0pt][l]{AffectGPT~\cite{lian2025affectgpt}} & {In the text, the caption reads: ``Don't ask why, do you want to eat duck?'' This phrase could be a query or invitation from one person to another. Considering the speaker's's tone described as high and the speaker possibly feeling anxious or tense, along with the visual clues of a girl's furrowed brow and possible discomfort, we can infer that this statement might carry an element of \textcolor{blue}{urgency or impatience}. Thus, it might come across as impatient or anxious, aligning with the girl's displayed sense of \textcolor{red}{uncertainty or concern.} }  \\
\hline
\makebox[0pt][l]{Emotion-LLaMA~\cite{cheng2024emotion}} & {In the video, a young girl's facial expressions and tone suggest a sense of annoyance or frustration her narrowed eyes and raised eyebrows indicate a heightened emotional state, while her negative tone and words, don't ask why, \textcolor{green}{do you want to eat for free?}, imply assertiveness and possibly anger therefore, it is inferred that his emotional state is \textcolor{blue}{angry} }  \\
\hline
Emotion-LLaMAv2 & {$<$think$>$The speaker's vocal tone is sharp and raised, with a fast speech rate and emphasis on ``Don't ask why'', indicating a strong emotional response the facial expressions, including a furrowed brow and tightened eyelids, align with feelings of anger or frustration. The visual scene, set in a dimly lit room with a serious expression, further supports this interpretation the context of the conversation, particularly the phrase ``do you want to eat Peking duck?'', suggests a confrontational or defensive stance. The combination of vocal clues, facial expressions, and contextual elements consistently points towards an emotional state of anger, as the speaker appears to be reacting to a perceived provocation or challenge. $<$/think$>$$<$answer$>$\textcolor{blue}{angry}$<$/answer$>$}  \\
\hline\thickhline
\end{tabular}}
% }
% \vspace{1mm}
% \vspace{-5mm}
\label{tab:reason2}
% \end{minipage}
\end{table*}

\begin{figure}
\setlength{\abovecaptionskip}{1pt}
\setlength{\belowcaptionskip}{1pt}
    \centering
    \includegraphics[width=0.95\linewidth]{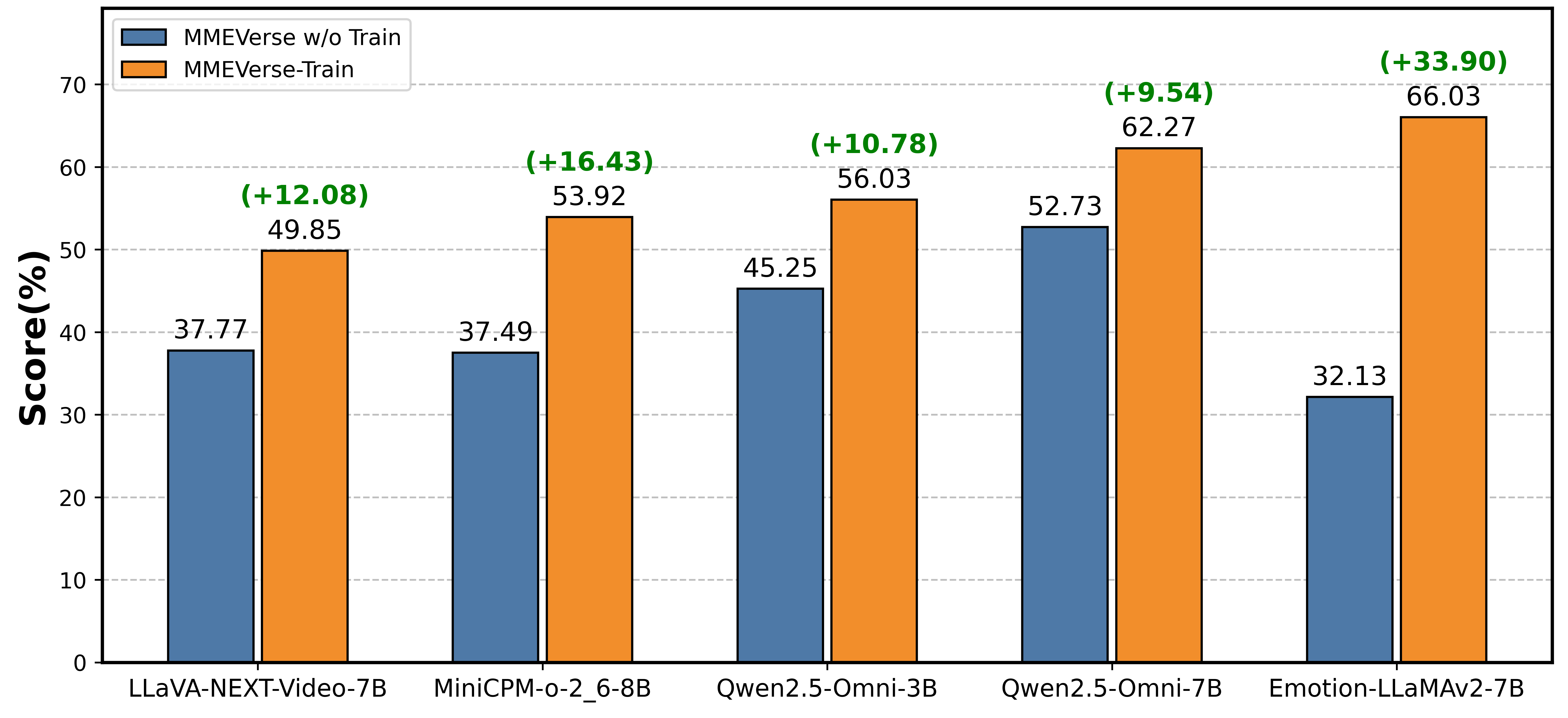}
    \caption{Ablation of \textbf{MLLM architecture.} Our Emotion-LLaMA-v2 is based on MiniGPT4v2, with LLaMA2 serving as its backbone. Qwen2.5-Omni and MiniCPM-o are natively supported audio, video, and text.}
    \label{fig:backbone}
    \vspace{-4mm}
\end{figure}

\begin{figure*}[t]
\setlength{\abovecaptionskip}{1pt}
\setlength{\belowcaptionskip}{1pt}
  \centering
  \includegraphics[width=1\linewidth]{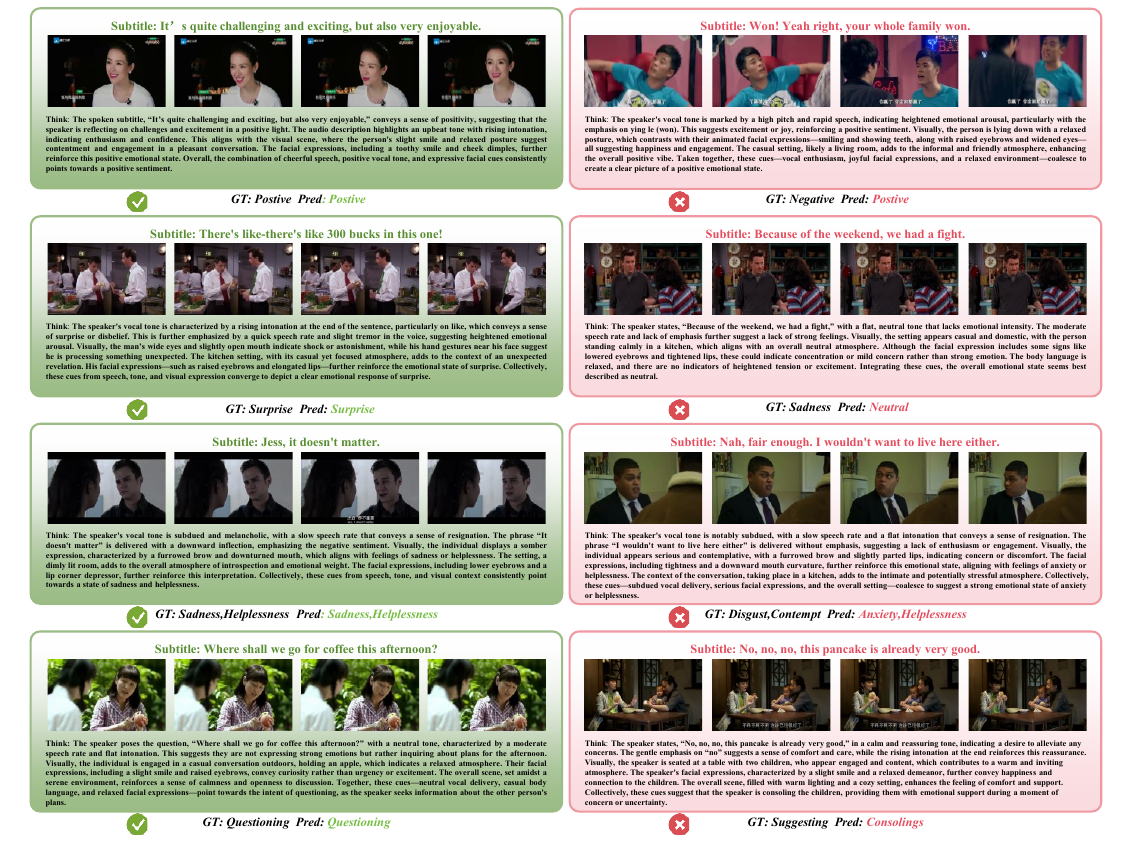}
  \caption{Visualization of our Emotion-LLaMAv2 on multimodal emotion understanding tasks. \pxj{The left side displays correct predictions with high confidence across various emotion tasks, while the right side shows error predictions.} }
  % \vspace{-4mm}
  \label{fig:logits}
\end{figure*}

\noindent\textbf{Ablation on Modality Encoders}. 
We conduct ablation studies to identify the optimal unimodal encoders and their multimodal combinations, with results shown in Table~\ref{tab:encFusionEval}. In the audio domain, HuBERT achieves the highest performance (58.92\%), outperforming Whisper by 2.64\%. For static image encoding, EVA proves superior (57.7\%), which we attribute to its pre-alignment with the LLaMA2 weights used in our base model MiniGPTv2. For temporal video encoding, applying strong image encoders (e.g., EVA, CLIP) to individual frames is more effective than dedicated video encoders such as VideoMAE, as this approach better preserves fine-grained visual clues critical for emotion recognition.

Building on these unimodal findings, we evaluate multimodal combinations. Interestingly, while HuBERT excels in unimodal audio tasks, the Whisper+EVA pairing yields the best audio-image performance (61.73\%), suggesting stronger feature complementarity between Whisper and EVA in multimodal contexts. The combination of Whisper for audio, EVA for global frames, and EVA for temporal frames achieves the highest overall score, forming the basis of our Emotion-LLaMAv2 model. \pxj{Our preliminary Emotion-LLaMA utilizes two encoders, MAE and VideoMAE, for facial sequences with individually pooled tokens, along with a pooled audio representation using HuBERT. This configuration results in performance that is significantly inferior to other trimodal approaches due to its inability to preserve temporal emotional clues}. The Emotion-LLaMAv2 not only surpasses our preliminary Emotion-LLaMA but also establishes a streamlined end-to-end pipeline by removing the need for separate face-extraction preprocessing~\cite{ramesh2022leveraging}.

% \noindent \textbf{Ablation on Global Visual Representation}. Table \ref{tab:ablation_frame} reveals the impact of the global visual representation choice. We first evaluate three practical, heuristic-based strategies: selecting the first, middle, or a random frame. The results show that the selection method is critical to performance. Among these, the 'Middle' frame strategy is the most effective, achieving a score of 62.46 on our primary benchmark, MMEVerse-Bench, which suggests that the central portion of a clip is generally more likely to contain salient emotional information.To probe the full potential of this global visual representation, we then introduce an oracle strategy by selecting the 'Emotion Peak' frame, which is identified using ground-truth labels. This oracle setting boosts performance significantly to 63.37.

% The findings from this ablation are twofold. For practical implementation, selecting the 'Middle' frame is a robust and effective heuristic. Furthermore, the substantial 0.91-point performance gap between the best heuristic and the oracle setting demonstrates that the **expressive content of the chosen frame is far more critical than its temporal location**. This highlights that the full potential of our model could be unlocked by more accurately identifying and leveraging these information-dense "peak" moments.

\noindent\textbf{Ablation on Temporal Representations}. 
Figure~\ref{fig:TemporalRepresentations} evaluates frame sampling for video and audio clues. In each case, one modality is fixed while the other varied. From Figure~\ref{fig:TemporalRepresentations}(a), increasing audio frames generally improves performance, saturating at 64. For video frames in Figure~\ref{fig:TemporalRepresentations}(b), performance peaks at 16 with a score of 66.03\%, but drops sharply when sampling 32 or 64. This decline can be explained by two factors: (i) emotional videos are typically short and exhibit consistent facial expressions, reducing the need for extensive sampling; (ii) sampling too few or too many tokens in one modality disrupts cross-modal balance, negatively affecting performance.  

We also evaluate the spatial pooling size of the temporal encoder, set by default to 2$\times$2, yielding 64 tokens. As shown in Figure~\ref{fig:TemporalRepresentations}(c), maintaining highly detailed spatial embeddings (e.g., 8$\times$8 grid with 64 tokens per frame) decreases performance and increases cost. The default 2$\times$2 granularity achieves the best results, striking a balance between preserving sufficient spatial detail and enabling efficient temporal encoding, thereby capturing critical spatio-temporal information for optimal performance.

\noindent\textbf{The impact of MLLM Architecture}. 
To identify the most effective base architecture for emotion recognition, we evaluated several recent MLLMs with their default configurations, including LLaVA-NEXT-Video-7B, MiniCPM-o-2.6-8B, Qwen2.5-Omni-3B, and Qwen2.5-Omni-7B. For the Omni models, decoder components were excluded during training and evaluation. 

Figure~\ref{fig:backbone} compares model performance with and without instruction tuning on the MMEVerse dataset. Several key findings emerge: First, all tested models show substantial improvement after tuning with MMEVerse-Train, underscoring the dataset’s effectiveness in enhancing multimodal emotion recognition. Second, Emotion-LLaMAv2, built upon MiniGPT4v2 and augmented with pre-fusion, temporal visual, and audio information, achieves the highest overall score of 66.05\%, establishing it as the most effective architecture. Third, for Qwen2.5-Omni models, stronger zero-shot performance correlates with better post-training results; notably, Qwen2.5-Omni-7B improves from 52.73\% to 62.27\%, ranking second overall. Fourth, LLaVA-NEXT-Video-7B shows limited gains, reaching 49.85\% after tuning, likely constrained by its reliance on video data and lack of audio clues.

\subsection{Qualitative Analysis and Visualization}

\noindent\textbf{Qualitative Analysis of Emotion Reason}. 
To illustrate the qualitative performance of Emotion-LLaMAv2, we present a comparison of emotion reasoning results across models. Table~\ref{tab:reason2} reports the reasoning outputs of several mainstream models. The video depicts a person smiling while questioning another individual, an expression of dissatisfaction suggesting an angry emotional state. Accurate reasoning for this sample requires integrating information from multiple modalities. PandaGPT and Valley capture the correct visual features but fail to incorporate other modalities, misclassifying the emotion as happy. Emotion-LLaMA reaches the correct inference, but its reasoning is compromised by hallucinations. AffectGPT provides a reasonable description yet misclassifies the emotion category. In contrast, Emotion-LLaMAv2 demonstrates a comprehensive understanding by recognizing the speaker’s tone and combining subtle facial expressions with multimodal information for accurate reasoning. This example highlights the superiority of our model in integrating emotional clues across modalities, yielding precise and contextually relevant recognition.

\noindent\textbf{Case Study of Emotion-LLaMAv2}. 
Figure~\ref{fig:logits} presents test examples using Emotion-LLaMAv2 on tasks including sentiment analysis, emotion recognition, multi-label recognition, and intent recognition. From correct prediction cases (left), we observe that Emotion-LLaMAv2 effectively captures tone and subtle facial expressions, correctly categorizing instances when audio, video, and text clues are consistent. 
\pxj{However, in the examples on the right side of Figure~\ref{fig:logits}, the model can be misled by sarcasm or conflicting expressions that convey implicit emotions. For instance, when a speaker says, ``Won! Yeah right, your whole family won'' in an excited tone, the model recognizes it as a positive sentiment, while the ground truth is negative due to the sarcastic expression.}

\section{Conclusion and discussion}
\label{sec:conclusion}
In this paper, we introduced an advanced multimodal emotion large model, Emotion-LLaMAv2, designed to recognize and understand emotion states in the real world. Compared to our preliminary Emotion-LLaMA, Emotion-LLaMAv2 is an end-to-end unified emotion framework and incorporates a sophisticated Conv-Attention pre-fusion module and a perception-to-cognition curriculum training strategy. Additionally, to support large-scale instruction-tuning and reproducible benchmarking, we curated the MultiModal Emotion uniVerse (MMEVerse) dataset, which aggregates twelve existing datasets (e.g., IEMOCAP, MELD, DFEW, MAFW) into a unified format and re-annotates them through a multi-agent pipeline integrating Qwen2-Audio, Qwen2.5-VL, and GPT-4o.  Extensive experiments demonstrated the effectiveness of our Emotion-LLaMAv2 and MMEVerse. We believe that Emotion-LLaMAv2 has the potential to advance the field of multimodal emotion understanding and pave the way for more empathetic and emotionally intelligent AI systems.

\noindent\textbf{Limitation and discussion.} 
Despite the advancements presented in Emotion-LLaMAv2, several limitations remain. First, while our model performs well on various emotion recognition tasks, it may struggle with emotional reversal and sarcasm. This is particularly relevant for emotions that are culturally specific or context-dependent, where the model's accuracy might diminish. Second, the model’s performance can be affected by the quality of the input data. For example, noise in audio or video streams can impair the model's ability to extract relevant emotional clues, leading to misclassification. Additionally, while Emotion-LLaMAv2 demonstrates improved integration of multimodal information, the underlying mechanisms of how different modalities interact are not fully transparent.
Lastly, the scalability of Emotion-LLaMAv2 to real-time applications remains a question. The complexity of processing multiple modalities simultaneously may introduce latency issues, which could hinder user experience in practical applications such as conversational AI or emotion-aware systems.
% Future work should address these limitations by expanding the dataset to include a broader range of emotional expressions, enhancing preprocessing techniques to improve input quality, and developing methods for clearer interpretability of the model's outputs.
% \noindent \textbf{Ethics.} \czb{In this work, all datasets used are covered by signed usage agreements, ensuring they are solely used for academic research. The MERR dataset employed in our study is derived from the MER2023 dataset, which contains over 70,000 unannotated samples from movies and TV series. We have signed the necessary End User License Agreements (EULA) and obtained permission from the original data providers to use this data. Additionally, the open-source MERR dataset consists of emotion description JSON files but does not include the source videos. Researchers wishing to access the dataset must apply directly to the original data providers and sign the EULA to meet the relevant requirements. We ensure that the MERR dataset only includes content related to multimodal emotion descriptions and does not involve any discriminatory or biased material.}

% \section{Acknowledgments}
% This work was supported by the National Natural Science Foundation of China (62176165), the Stable Support Projects for Shenzhen Higher Education Institutions (20220718110918001), and the Natural Science Foundation of Top Talent at SZTU (GDRC202131), the support of NSF CISE under grant number 1937998.

% \newpage

\bibliographystyle{IEEEtran}
\normalem
\bibliography{reference.bib}{}
\appendix

\section{MMEVerse details}
\subsection{Annotation pipeline}
\label{sup:annot}
\begin{figure}[t]
\centering
    \includegraphics[width=\linewidth]{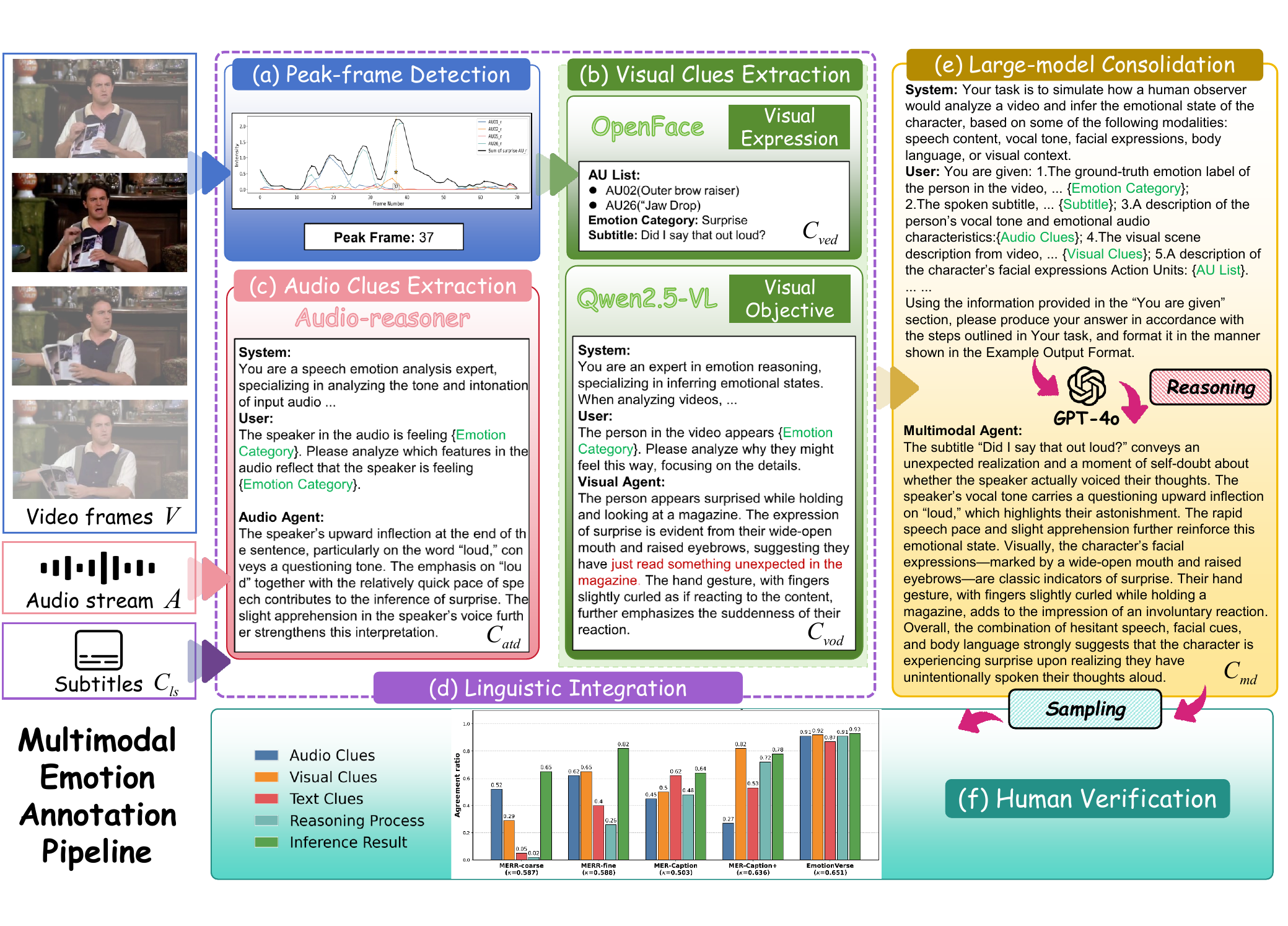}
    \caption{\textbf{Multimodal Emotion annotation pipeline} includes a)~peak-frame detection, b)~visual clues extraction, c)~audio clues extraction, d)~linguistic integration, e)~GPT-4o consolidation, and f)~human verification, yielding multimodal emotional descriptions.}
\label{fig:dataAnnot}
\end{figure}

Figure~\ref{fig:dataAnnot} illustrates the complete pipeline of our annotation procedure. To enable large-scale annotation while ensuring semantic consistency across diverse data sources, we have developed a unified multimodal annotation framework. This framework systematically re-annotates each sample through structured analyses of visual, acoustic, and linguistic data.

For each video, we first extract Facial Action Units (AUs) from all frames using OpenFace. We then select the frame with the highest aggregated AU intensity score as the peak emotional frame, which serves as the anchor for subsequent visual expression analysis. This peak frame undergoes processing by two complementary vision models: OpenFace generates AU-based facial expression descriptors to capture fine-grained muscle movements, while Qwen2.5-VL-72B provides high-level contextual insights regarding scene layout, objects, and interactions. The outputs of these models create a visual expression description and a visual objective description, respectively.

Simultaneously, the audio stream is analyzed using Qwen2-Audio or Audio-Reasoner to extract prosodic and paralinguistic features, such as pitch variation, speaking rate, intensity, pauses, and hesitations, resulting in an audio tone description. Lexical subtitles further contribute semantic and pragmatic context that grounds emotional expressions within the spoken content. All unimodal cues are then integrated to form a preliminary multimodal emotional description, which is refined using GPT-4o.

To evaluate annotation reliability, a subset of samples undergoes assessment through a human user study, providing quantitative evidence of annotation quality and consistency.

\begin{figure}
    \centering
    \centering
    \includegraphics[width=0.8\linewidth]{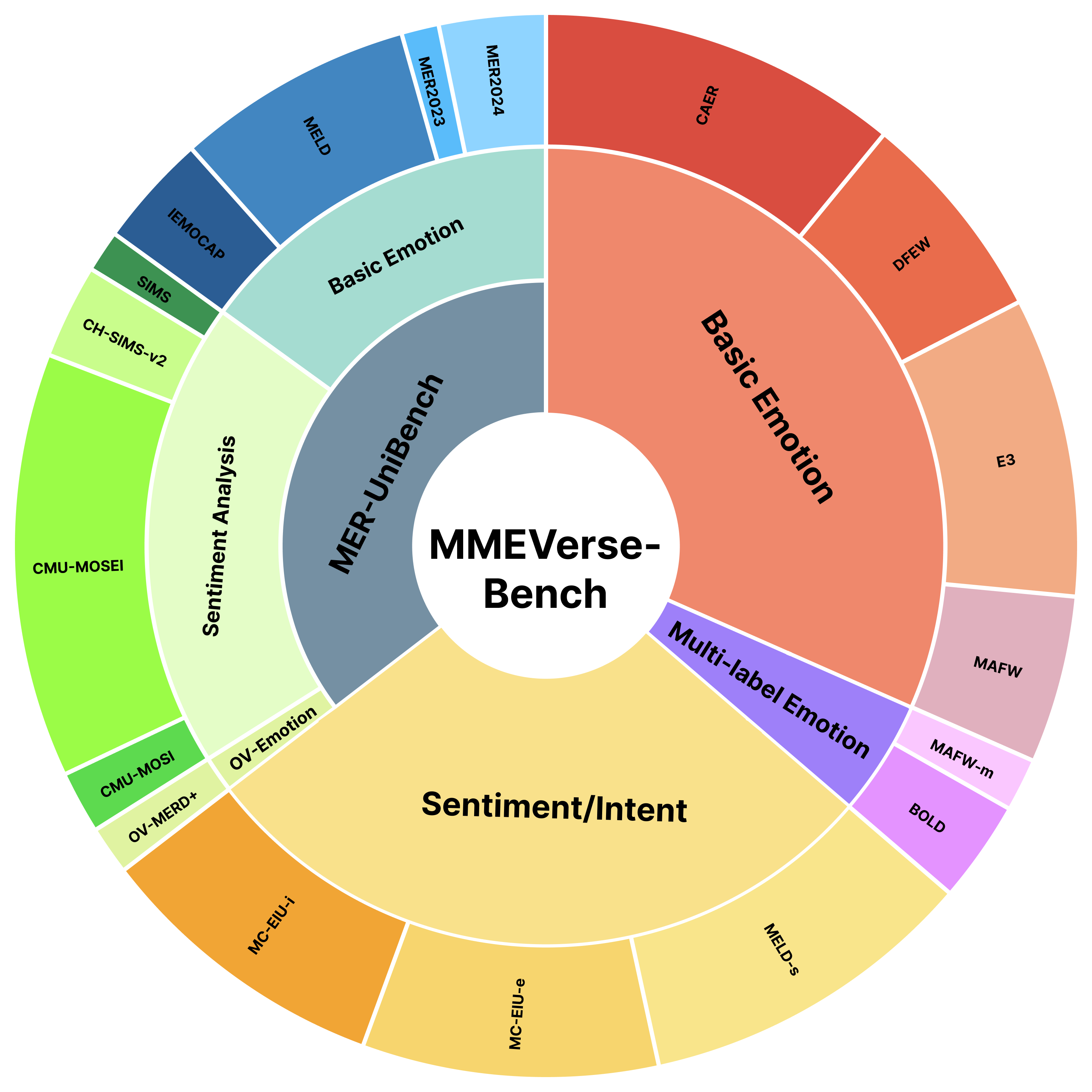}
    \caption{\textbf{Distribution comparison} between the proposed MMEVerse-Bench and the MER-UniBench~\cite{lian2025affectgpt}. MMEVerse-Bench provides a substantially larger and more balanced evaluation set.  }
\label{fig:BenchVS}
\end{figure}
   
\subsection{MMEVerse-Bench Visualization}
\label{sup:data}
Our MMEVerse-Bench consists of 18 subsets, including data of basic emotion, sentiment, intent, multi-label emotion, and OV-emotion, as illustrated in Figure~\ref{fig:BenchVS}. Compared to MER-UniBench, MMEVerse-Bench exhibits a more balanced evaluation distribution across cultural, situational, and multimodal conditions, enabling more rigorous and domain-agnostic assessments. Taken together, these properties position MMEVerse as a unified, scalable, and semantically aligned resource for next-generation multimodal emotion understanding.

\section{Training and Implementation Details}
\subsection{Instructions for Multimodal Emotion Recognition}
\label{sup:insRec}
\textbf{Box I} presents the instructions for training multimodal emotion recognition, i.e., the perception training phase.

\begin{figure}[t]
\begin{tcolorbox}[valign=bottom, colback=white, title=Box I: Prompts for Emotion Alignment Instruction Tuning]
\label{box1}
\begin{itemize}[leftmargin=2mm]
    \item ``Please determine which emotion label in the video represents: $\bigstar$.''
    \item ``Identify the displayed emotion in the video: is it $\bigstar$?''
    \item ``Determine the emotional state shown in the video, choosing from $\bigstar$.''
    \item ``Please ascertain the specific emotion portrayed in the video, whether it be $\bigstar$.''
    \item ``Assess and label the emotion evident in the video: could it be $\bigstar$?''
    \item ``Given the input video and audio, your task is to identify the emotion expressed by the person or people in the video. Your output must be only one emotion label strictly chosen from the following list:$\bigstar$.''
    \item ``Please classify the observed emotional state in the video using one of the following:$\bigstar$''
    \item ``Based on the visual and audio content of the video, identify the emotion expressed by the person. Choose one label from:$\bigstar$''
    \item ``Analyze the video and assign one of the following emotion labels to the person depicted:$\bigstar$''
    \item ``After watching the video, decide which of the following emotions is being expressed:$\bigstar$''
\end{itemize}
$\bigstar$: Emotion categories within the original dataset.
\end{tcolorbox}
\end{figure}

\subsection{Instructions for Multimodal Emotion Reasoning}
\label{sup:insrea}
\textbf{Box II} presents the instructions used in training multimodal emotion reasoning, i.e., the perception and cognition training phase.
% {\small
\begin{figure*}[t]
  \centering
\begin{tcolorbox}[colback=white, title=Box II: Prompts for Joint Recognition and Reasoning]
\begin{itemize}[leftmargin=2mm]            
    \item ``The possible emotions are: $\bigstar$. Based on what you see and hear in the video, including facial expressions, gestures, vocal tone, and spoken words, identify the emotion the person is expressing and explain which clues led to your conclusion.''
    \item ``Choose one emotion from the following list: $\bigstar$. Watch the video and use both visual signals, such as facial expressions and body movements, and auditory signals, such as tone and intonation, to infer the person’s emotional state. Please describe your reasoning process clearly.''
    \item ``You are given a video containing both visual and audio information. The possible emotion categories are: $\bigstar$. First, analyze the video by reasoning through facial expressions, gestures, tone of voice, and spoken content. Write your reasoning inside the <think> and </think> tags. Then select the most appropriate emotion and place it inside the <answer> and </answer> tags.''
    \item ``Watch the video and consider all visual and auditory clues, including facial expressions, body movements, voice pitch, tempo, and speech content. The emotion must be one of: $\bigstar$. Use <think> to explain your reasoning step by step, and then provide the final emotion label in <answer>.''
    \item ``Analyze the multimodal signals in the video. The goal is to infer the person’s emotional state from the following options: $\bigstar$. In the <think> section, describe how the visual and audio evidence supports your reasoning. Then output the most likely emotion label in <answer>.''   
\end{itemize}
\end{tcolorbox}
\end{figure*}
% }

\subsection{Prompts for emotion reasoning evaluation}
\label{sup:reaEval}
\textbf{Box III} presents the prompt used in emotion reasoning evaluation.
% {\small

\begin{figure*}[t]
% \vspace{-10pt}
\begin{tcolorbox}[valign=top, colback=white, title=Box III: Prompts for emotion reasoning evaluation.]
Below, the “Actual Description” and “Predicted Description” of a character are given. Please follow the steps to calculate
the score for the “Predicted Description”. The score should range
from 1 to 10. In the end, only output the numerical value of the
predicted score along with the reasoning.
\begin{itemize}[leftmargin=2mm]
\item[1] Summarize the emotional state description of the character
from the “Actual Description”.
\item[2] Summarize the emotional state description of the character
from the “Predicted Description”.
\item[3] Calculate the overlap between the “Predicted Description”
and the “Actual Description”. The higher the overlap, the
higher the score.
\item[4] Output format: ’Predicted Score’: Predicted Score; ’Reason’:
Reason
\end{itemize}
Input:\\
“Actual Description”: <description annotation>\\
“Predicted Description”: <predicted description>\\
Output:
\end{tcolorbox}
\end{figure*}

\end{document}